\pdfoutput=1
\documentclass[11pt]{article}
\usepackage{authblk}
\usepackage[final]{acl2023}
\usepackage{times}
\usepackage{graphicx}
\usepackage{latexsym}
\usepackage{array}
\usepackage{booktabs}
\usepackage{pgfplots}
\usepackage{algorithm}
\usepackage{algpseudocode}
\usepackage{graphicx}
\usepackage[T1]{fontenc}
\usepackage[utf8]{inputenc}
\usepackage{microtype}
\usepackage{xcolor}
\usepackage{soul}
\usepackage{subcaption}
\usepackage{tikz}
\usepackage{pgfplots}
\usepackage{adjustbox}
\usepackage{times}
\usepackage{xcolor}
\usepackage{graphicx}
\usepackage{latexsym}
\usepackage{array}
\usepackage{booktabs}
\usepackage{pgfplots}
\usepackage{algorithm}
\usepackage{algpseudocode}
\usepackage{graphicx}
\usepackage[T1]{fontenc}
\usepackage[utf8]{inputenc}
\usepackage{microtype}
\usepackage{xcolor}
\usepackage{soul}
\usepackage{fullpage}
\usepackage{enumitem}
\usepackage{pgfplots}
\usepackage{algorithm}
\usepackage{algpseudocode}
\usepackage{graphicx}
\usepackage{placeins}
\usepackage{tabularx}
\usepackage{makecell}
\usepackage{booktabs}       
\usepackage{array}          
\usepackage{colortbl}
\newcolumntype{P}[1]{>{\centering\arraybackslash}p{#1}}
\newcolumntype{M}[1]{>{\centering\arraybackslash}m{#1}}
\newcommand{\greyrule}{\arrayrulecolor{black!30}\midrule\arrayrulecolor{black}}
\usepackage{amsmath}
\usepackage{mathtools}
\title{Quick Dense Retrievers Consume KALE: Post Training Kullback–Leibler Alignment of Embeddings for Asymmetrical dual encoders}
\author[1]{Daniel Campos \thanks{~~~Corresponding author: dcampos3@illinois.edu}}
\author[2]{Alessandro Magnani}
\author[1]{ChengXiang Zhai}
\affil[1]{Department of Computer Science, the University of Illinois Urbana-Champaign}
\affil[2]{Walmart Labs}
\begin{document}
\maketitle
\begin{abstract}
In this paper, we consider the problem of improving the inference latency of language model-based dense retrieval systems by introducing structural compression and model size asymmetry between the context and query encoders. First, we investigate the impact of pre and post-training compression on the MSMARCO, Natural Questions, TriviaQA, SQUAD, and SCIFACT, finding that asymmetry in the dual-encoders in dense retrieval can lead to improved inference efficiency. Knowing this, we introduce \emph{Kullback–Leibler Alignment of Embeddings} (KALE), an efficient and accurate method for increasing the inference efficiency of dense retrieval methods by pruning and aligning the query encoder after training. Specifically, KALE extends traditional Knowledge Distillation after bi-encoder training, allowing for effective query encoder compression without full retraining or index generation. Using KALE and asymmetric training, we can generate models which exceed the performance of DistilBERT despite having 3x faster inference. 
\end{abstract}
\section{Introduction}
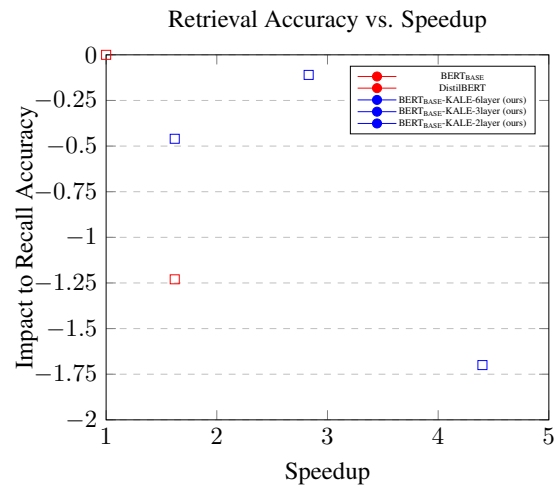
\begin{figure}[!htb]
\begin{tikzpicture}
\scalebox{0.85}{
\begin{axis}[
    title={Retrieval Accuracy vs. Speedup},
    ylabel={Impact to Recall Accuracy},
    xlabel={Speedup},
    ymin=-2, ymax=0,
    xmin=1 , xmax=5,
    ytick={0,-0.25,-0.5,-0.75, -1,-1.25, -1.5, -1.75,-2},
    xtick={1, 2, 3, 4,5},
    legend pos=north east,
    ymajorgrids=true,
    grid style=dashed,
    legend style={nodes={scale=0.4, transform shape}}, 
    legend image post style={mark=*}
]
\addplot[
    color=red,
    mark=square,
    ]
    coordinates {
    (1,0)
    };

\addplot[
    color=red,
    mark=square,
    ]
    coordinates {
    (1.62,-1.23)
    };
\addplot[
    color=blue,
    mark=square,
    ]
    coordinates {
    (1.62,-0.46)
    };
\addplot[
    color=blue,
    mark=square,
    ]
    coordinates {
    (2.83,-0.11)
    };
\addplot[
    color=blue,
    mark=square,
    ]
    coordinates {
    (4.4,-1.7)
    };

\legend{BERT\textsubscript{BASE}, DistilBERT,  BERT\textsubscript{BASE}-KALE-6layer (ours),BERT\textsubscript{BASE}-KALE-3layer (ours), BERT\textsubscript{BASE}-KALE-2layer (ours) }
 \end{axis}}
\end{tikzpicture}
    \centering
    \caption{Using KALE and asymmetric training on the lead to when measuring QPS vs. Recall at 100 on the NQ dataset.  Using Asymmetry and KALE, it is possible to 3x QPS with nearly no loss in accuracy and 4.5x with under 2\% loss in accuracy. We calculate QPS as the mean number of queries per second with a batch size of 1 and a max sequence length of 32 on a T4 GPU. Impact on retrieval accuracy is measured by the relative drop in retrieval accuracy at 100}
    \label{fig:speed}
\end{figure}
A bi-encoder-based retrieval, often called dense retrieval, is a retrieval function that leverages the vector representation of queries and documents as a proxy for relevance. Using two encoders, one for the query and one for the document, the input data is mapped into a common latent space where closeness becomes a proxy for relevance. \\
Dense retrievers have become increasingly popular due to their ability to capture the semantic relationships between query and document terms. However, bi-encoder-based models can also be computationally expensive, particularly when dealing with large datasets. As a result, there has been a growing interest in methods for compressing these models to reduce their computational complexity without sacrificing performance.\\
While the use of smaller models \cite{wang2020minilm} has provided a path to improving model performance, compression cannot be adjusted to suit varying latency needs. In other words, a model must match latency requirements before it can be experimented with. Additionally, since bi-encoders require a complete index generation to evaluate performance iteratively compressing models and retraining them can be very expensive. Seeing the bottleneck caused by trying to train compressed models for retrieval we explore approaches to compress models after training. By doing so it becomes cheaper to evaluate the impact of compression of retrieval and generate variants of many sizes. \\
In this paper, we explore the role of asymmetry in the size of query and document encoders that leverage language models. Through experiments on several benchmarks, we demonstrate that our approach can significantly reduce the number of parameters in the bi-encoder model without sacrificing performance. \\
As shown in figure \ref{fig:speed}, the combination of asymmetric bi-encoders and post-training KALE allows for 3x more QPS than an uncompressed bi-encoder with less than 1\% loss in accuracy and nearly 5x with less than 2\%.    \\
Building on the favorable implications of asymmetry for efficient inference, we introduce a compression mechanism called \textbf{K}ullback-Leibler \textbf{Al}lingment of \textbf{E}mbeddings (KALE). KALE uses an alignment of representations to compress models without requiring any form of retraining or index regeneration. \\
To ground our approaches, we evaluate the effectiveness of KALE and asymmetry on several benchmark datasets and compare the results to existing efficient inference approaches. \\
The following research questions drive our work: 
\begin{itemize}
    \item Is the performance of dense retrieval methods more driven by the query or document encoder size?
    \item Is it possible to compress query encoders without retraining and index regeneration?
    \item How can dense retrieval asymmetry and post-training alignment be leveraged to improve query encoder latency?
\end{itemize}
It is in answering these questions that we deliver the following contributions: 
\begin{itemize}
\item We present the first robust studies on the role of document-query encoder symmetry, demonstrating that the size of the document encoder dominates performance. 
\item We introduce and demonstrate the effectiveness of KALE, a post-training compression and alignment approach demonstrating its effectiveness and 
\item We empirically demonstrate on various benchmarks how Asymmetric Compression can lead to 4.5 better QPS with 1\% loss in recall accuracy at 100.
\end{itemize}
\section{Related Work}
\textbf{Transformer Based Language Models} such as BERT \cite{Devlin2019BERTPO} provide contextual language representations built on the Transformer architecture \cite{Vaswani2017AttentionIA} which can be specialized and adapted for specific tasks and domains \cite{Lee2020BioBERTAP}. Using contextual word representations, it becomes relatively easy to excel at a broad range of natural language processing tasks such as Question Answering, Text Classification, and sentiment analysis. \\
\textbf{Bi-Encoders}, commonly called dual-encoders or dense retrievers, decompose ranking by leveraging the inner product of query and document representations to produce a relevance score for query document pairs. While not as accurate at cross-encoders \cite{Reimers2019SentenceBERTSE}, they are more efficient for inference and easier to deploy. Bi-encoder document representations are query invariant, allowing them to be pre-computed and loaded into an Approximate Nearest Neighbor (ANN) such as FAISS \cite{johnson2019billion}. \\
At runtime, a query is an encoder into a latent space, and the $k$ documents are retrieved using a nearest neighbor algorithm such as HNSW \cite{Malkov2016EfficientAR}. Since the entire document index has already been created the retrieval latency is limited to a single call of the query encoder. \\
Bi-encoders commonly leverage LLM such as BERT \cite{Devlin2019BERTPO} to retrieve short passages of text leading to the task descriptor of Dense Passage Retrievers (DPR) \cite{Karpukhin2020DensePR}. Driven by their efficiency in deployment and relevance performance, DPR-based models have rapidly become the building blocks for systems doing product search \cite{Magnani2022SemanticRA}, open domain question answering \cite{Karpukhin2020DensePR} and customer support \cite{Mesquita2022DenseTR}.\\
\textbf{Efficient Inference} study methods and models which decrease the model execution cost while minimizing the losses to model performance. \\
Knowledge Distillation \cite{Hinton2015DistillingTK} is a training method where a model, called the \textit{student}, learns to emulate a \textit{teacher} model, which is commonly larger or better performing than the \textit{student}.\\
Unstructured pruning removes individual weights or groups of weights in a model by applying a mask or setting the weight values to 0. When paired with a sparsity-aware inference engine, it is possible to gain 3-5x speedups in inference throughput with little to no loss in accuracy \cite{Kurti2022TheOB}. \\
Structured pruning removes fundamental structural components in a language model, such as individual attention heads \cite{Voita2019AnalyzingMS} or entire model layers \cite{sanh2019distilbert}. Removing entire model layers is one of the most pervasive approaches, as latency gains are easy to realize, and pruning is straightforward. \\
While their training regimes may differ, models like DistilBERT \cite{sanh2019distilbert} and TinyBERT \cite{Jiao2020TinyBERTDB}, and MiniLM \cite{wang2020minilm} leverage structural pruning as ways of generation 2-10x speedups. \\
Methods like quantization \cite{Pouransari2020LeastSB} \cite{Zafrir2019Q8BERTQ8}, early exiting \cite{Xin2020DeeBERTDE} or token pruning \cite{Kim2021LearnedTP} have been effective in other NLP tasks. Still, our work primarily focuses on structured pruning and its relationship with asymmetry. We leave studying the impacts of asymmetry on these compression methods to future work.  \\
\textbf{Asymmetrical deep learning} broadly refers to any non-uniformity in shape or attribute of models. Traditional modeling approaches favor uniformity as it is preferable for optimization algorithms \cite{Mihaylova2019ScheduledSF}, and using models for inference should match training as closely as possible \cite{Ranzato2015SequenceLT} as improvements in training loss during optimization result in improvements in model performance during inference. However, this does not account for cost or latency asymmetries during usage. Kasai et al. demonstrated how the sequence-to-sequence encoder depth dominates language model performance for machine translation \cite{Kasai2020DeepES}. Tay et al. 2021 extend this work by finding a \textit{DeepNarrow} which shows that for broad language modeling, it is possible to have 50\% fewer parameters and a 40\% faster inference with no loss in accuracy.\\
\textbf{Embedding Distillation}
Concurrent to our work on bi-encoder compression, Kim et al. 2023 study how distillation in embeddings leads to general compression of bi-encoders and cross-encoders \cite{Kim2023EmbedDistillAG}. Our work differs from theirs as we focus on the role of asymmetry between query and document encoders and how to leverage it for improved inference efficiency.

\section{Method}
The use of representation models for retrieval begins with a document space $d$ and a query space $q$ where each of which is generated by some model $m$. Models do not need to share the same initialization, shape, or size, but their representation vectors must share size without some projection. These two models learn a notion of relevance by training to minimize the distance of positive query-document pairs as shown in equation \ref{eq:dis} where $\textbf{x}$ is a query vector and $\textbf{y}$ is a document vector, and $\cdot$ denotes the dot product of the vectors.\\
\begin{equation}
L = 1 - \frac{\textbf{x} \cdot \textbf{y}}{|\textbf{x}||\textbf{y}|}
 \label{eq:dis}
\end{equation}
The query and document encoder models are commonly initialized with a pre-trained language model such as BERT. Then, using pairs of labels for positive relevance scores for queries and documents, the models are trained to minimize the distance between queries and their relevant documents \cite{Karpukhin2020DensePR} \\
\begin{figure*}[!htb]
\begin{tikzpicture}
\scalebox{1}{
\begin{axis}[
    title={Encoder layers Vs. Impact on Retrieval Accuracy   },
    xlabel={Query Encoder Layers},
    ylabel={Impact to Retrieval Accuracy},
    xmin=0, xmax=13,
    x dir=reverse,
    ymin=-35 , ymax=0,
    xtick={1,2,3,6,9,12},
    ytick={0, -10,-20,-30},
    legend pos=north east,
    ymajorgrids=true,
    grid style=dashed,
    legend style={nodes={scale=0.4, transform shape}}, 
    legend image post style={mark=*}
]
\addplot[
    color=red,
    mark=square,
    ]
    coordinates {
    (12, 0) (9, -7.00) (6,-7.8) (3,-12.55) (2, -12.76) (1, -26.12)
    };

\addplot[
    color=blue,
    mark=square,
    ]
    coordinates {
    (12, -7.07) (9, -8.08) (6,-10.3) (3,-14.98) (2, -15.92) (1, -29.83)

    };

\addplot[
    color=green,
    mark=square,
    ]
    coordinates {
    (12, -9.57) (9, -10.33) (6,-10.23) (3,-16.2) (2, -17.2) (1, -25.46)
    };

\addplot[
    color=orange,
    mark=square,
    ]
    coordinates {
    (12, -11.9) (9, -14.01) (6,-14.95) (3,-16.2) (2, -20.74) (1, -25.46)
    };
\addplot[
    color=red,
    mark=square,
    ]
    coordinates {
    (12, -12.90) (9, -14.95) (6,-15.43) (3,-16.23) (2, -16.27) (1, -24.8)
    };

\addplot[
    color=yellow,
    mark=square,
    ]
    coordinates {
    (12, -28.13) (9, -27.99) (6,-28.3) (3,-28.58) (2, -29.17) (1, -31.18)
    };
\legend{12 document layers, 9 document layers, 6 document layers, 3 document layers, 2 document layers, 1 document layer }
 \end{axis}}

\end{tikzpicture}
    \centering
    \caption{Measuring the impact on recall at 20 on the NQ retrieval dataset by varying the number of transformer layers for the query encoder and document encoder \label{fig:asm-1}}
\end{figure*}
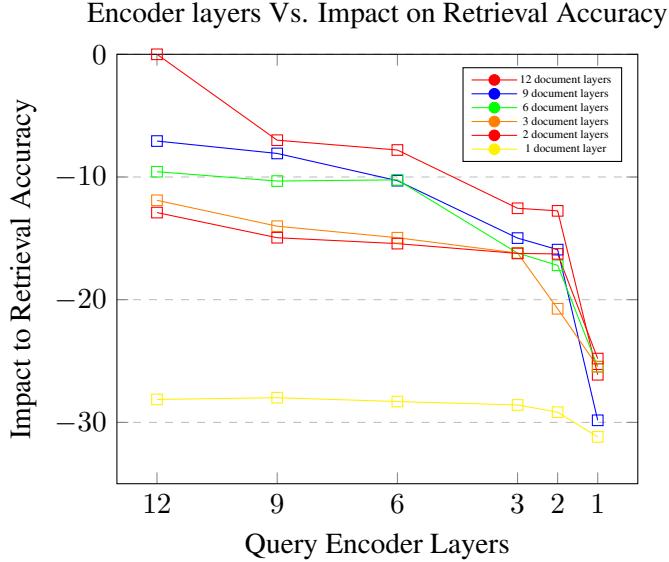
While it is common practice to initialize the query encoder and document encoder with identical language models, this ignores the cost asymmetry of the usage patterns. The document encoder is usually only used once during a large-scale batch generation of the index. Index generation happens in a latency-insensitive environment and can easily leverage many GPUs and large batch sizes to improve efficiency.\\
The query encoder runs every time a user issues a query, which can be irregular and sporadically. The query encoder responds to each user query independently. Thus, query encoders often use a batch size of 1 and commonly leverage small inference-optimized hardware like the T4 GPU or small CPUs. \\
Since the document encoder does not run very often, any improvement in latency produces a single fixed gain utterly dependent on the corpus size and index refresh cycle. The query encoder's user-facing nature means latency improvements occur whenever a user queries. 
\subsection{Role of model symmetry with Bi-encoders}
Since the query encoder runs many times online and the document encoder runs once, offline, we question: Is there some form of asymmetry between the query encoder and the document encoder that can be exploited? Do the two encoders need to be compressed symmetrically? \\
To answer this question, we explore the impact on the performance of pruning the query and document encoders on the NQ passage retrieval dataset \cite{Kwiatkowski2019NaturalQA}. Using a BERT-base uncased model with 12 transformer encoder layers, we generate structurally pruned models with 9,6,3,2 and 1 layer. We also further pre-train the three and six-layer models using knowledge distillation, represented as $6_{KD}$ and $3_{KD}$, from a 12-layer model on the Wikipedia-book corpus similar to distilBERT \cite{sanh2019distilbert}. \\
Then, using each of these seven models, we train dense retrieval models on the NQ passage retrieval dataset with variations of query and document models resulting in 72 variants. With each of these models, we generate a full index and evaluate retrieval performance on the development portion of the dataset. We do not tune any parameters to avoid overfitting and to explore asymmetry without overoptimizing. Each model's retrieval accuracy is evaluated with retrieval sets of depth 20, 100, and 200. We compare the impact of varying the encoders to the uncompressed baseline and a distilBERT model (denoted by $6_{db}$). \\
\begin{table}[!htb]
    \centering
    \caption{Impact of Structural pruning before fine-tuning on Retrieval Accuracy on NQ passage retrieval dataset}
    \scalebox{0.6}{\begin{tabular}{|l|l|l|l|l|l|l|}
    \hline        Layers $enc$ & Top 20 & Impact & Top 100 & Impact & Top 200 & Impact \\ \hline
        12 & 79.86\% & 0.00\% & 85.84\% & 0.00\% & 88.42\% & 0.00\% \\ \hline
        $6_{db}$ & 73.88\% & -7.49\% & 84.74\% & -1.29\% & 87.26\% & -1.31\% \\ \hline
        9 & 73.41\% & -8.08\% & 83.68\% & -2.51\% & 86.51\% & -2.16\% \\ \hline
        $6_{KD}$ & 75.04\% & -6.04\% & 85.15\% & -0.80\% & 87.45\% & -1.10\% \\ \hline
        6 & 71.69\% & -10.23\% & 83.30\% & -2.96\% & 86.04\% & -2.69\% \\ \hline
        $3_{KD}$ & 73.32\% & -8.19\% & 83.43\% & -2.80\% & 86.20\% & -2.51\% \\ \hline
        3 & 66.93\% & -16.20\% & 80.61\% & -6.09\% & 84.49\% & -4.45\% \\ \hline
        2 & 66.87\% & -16.27\% & 80.42\% & -6.32\% & 83.85\% & -5.17\% \\ \hline
        1 & 54.96\% & -31.18\% & 71.88\% & -16.26\% & 76.73\% & -13.22\% \\ \hline
    \end{tabular}}
    \label{tab:sym-nq}
\end{table}
Looking at the impact of symmetric compression as shown in table \ref{tab:sym-nq}, we see that the impact of compression is more pronounced with a small recall set as retrieval accuracy impact at 20 is ~3x that of at 200. As shown in table \ref{tab:sym-nq} we observe major accuracy gains by fine-tuning the pruned model with a ~4\% gap between $6$ and $6_{KD}$ and a ~8\% gap between $3$ and $3_{KD}$ with a 4\% gap for recall at 20 on the NQ dataset. \\
Looking at the impact of asymmetry of the depth of encoders as shown in table \ref{tab:asym-nq} and figure \ref{fig:asm-1} we find there is the size of the query and document encoders cause similar impacts on retrieval accuracy. A retriever with 3 layers in the query encoder and 12 in the document encoder loses 11.9\% of its retrieval accuracy and 12.55\% when the sizes of the document encoder and query encoders are flipped. These asymmetric retrievers perform better than the symmetric 3-layer models, which lose 16.2\% which highlights the ability to improve retrieval performance by having non-uniform compression.  \\
It is worth noting that having a larger document encoder is preferable to a larger query encoder which supports the notion that the document encoder is more important than the query encoder \cite{Li2021EncoderAO}.//
Similar results can be seen with the introduction of fine-tuned three and 6-layer models as shown in table \ref{tab:asym-nq-kd}. Unsurprisingly, KD-optimized language models outperform non-distilled models, and any asymmetrical variant that leverages a distilled model outperforms the un-distilled variant. Without further optimization, a model with a distilled 3-layer query encoder and a 12-layer document encoder will outperform a model with symmetrical 6-layer models despite being 2x faster. \\
\begin{table}[!htb]
    \centering
    \caption{Impact of Structural pruning before fine-tuning on Retrieval Accuracy on NQ passage retrieval dataset}
    \scalebox{0.5}{
    \begin{tabular}{|l|l|l|l|l|l|l|l|}
    \hline
        $layers_{q}$ & $layers_{d}$ & Top 20 & Impact & Top 100 & Impact & Top 200 & Impact \\ \hline
        12 & 12 & 79.86\% & 0.00\% & 85.84\% & 0.00\% & 88.42\% & 0.00\% \\ \hline
        \midrule

        9 & 12 & 74.27\% & -7.00\% & 84.40\% & -1.67\% & 86.95\% & -1.66\% \\ \hline
        6 & 12 & 73.63\% & -7.80\% & 84.27\% & -1.83\% & 86.79\% & -1.85\% \\ \hline
        3 & 12 & 69.83\% & -12.55\% & 82.58\% & -3.80\% & 85.35\% & -3.48\% \\ \hline
        2 & 12 & 69.67\% & -12.76\% & 82.19\% & -4.25\% & 84.68\% & -4.23\% \\ \hline
        1 & 12 & 59.00\% & -26.12\% & 75.37\% & -12.19\% & 81.00\% & -8.39\% \\ \hline
        \midrule
        12 & 9 & 74.21\% & -7.07\% & 84.40\% & -1.67\% & 87.06\% & -1.53\% \\ \hline
        9 & 9 & 73.41\% & -8.08\% & 83.68\% & -2.51\% & 86.51\% & -2.16\% \\ \hline
        6 & 9 & 71.63\% & -10.30\% & 83.05\% & -3.25\% & 85.98\% & -2.76\% \\ \hline
        3 & 9 & 67.89\% & -14.98\% & 80.94\% & -5.71\% & 84.79\% & -4.10\% \\ \hline
        2 & 9 & 67.15\% & -15.92\% & 80.53\% & -6.19\% & 83.66\% & -5.39\% \\ \hline
        1 & 9 & 56.04\% & -29.83\% & 73.35\% & -14.55\% & 78.12\% & -11.65\% \\ \hline
        \midrule
        12 & 6 & 72.22\% & -9.57\% & 83.41\% & -2.83\% & 85.84\% & -2.91\% \\ \hline
        9 & 6 & 71.61\% & -10.33\% & 83.30\% & -2.96\% & 85.93\% & -2.82\% \\ \hline
        6 & 6 & 71.69\% & -10.23\% & 83.30\% & -2.96\% & 86.04\% & -2.69\% \\ \hline
        3 & 6 & 66.93\% & -16.20\% & 80.28\% & -6.48\% & 83.96\% & -5.04\% \\ \hline
        2 & 6 & 66.12\% & -17.20\% & 80.33\% & -6.42\% & 83.49\% & -5.58\% \\ \hline
        1 & 6 & 59.53\% & -25.46\% & 75.37\% & -12.19\% & 79.83\% & -9.71\% \\ \hline
        \midrule
        12 & 3 & 70.36\% & -11.90\% & 81.72\% & -4.80\% & 84.60\% & -4.32\% \\ \hline
        9 & 3 & 68.67\% & -14.01\% & 80.47\% & -6.25\% & 84.46\% & -4.48\% \\ \hline
        6 & 3 & 67.92\% & -14.95\% & 80.06\% & -6.74\% & 83.85\% & -5.17\% \\ \hline
        3 & 3 & 66.93\% & -16.20\% & 80.61\% & -6.09\% & 84.49\% & -4.45\% \\ \hline
        2 & 3 & 63.30\% & -20.74\% & 78.37\% & -8.71\% & 83.02\% & -6.11\% \\ \hline
        1 & 3 & 59.53\% & -25.46\% & 75.68\% & -11.84\% & 80.08\% & -9.43\% \\ \hline
        \midrule
        12 & 2 & 69.56\% & -12.90\% & 81.33\% & -5.25\% & 84.49\% & -4.45\% \\ \hline
        9 & 2 & 67.92\% & -14.95\% & 80.75\% & -5.93\% & 84.32\% & -4.64\% \\ \hline
        6 & 2 & 67.53\% & -15.43\% & 80.33\% & -6.42\% & 83.82\% & -5.20\% \\ \hline
        3 & 2 & 66.90\% & -16.23\% & 80.36\% & -6.38\% & 84.24\% & -4.73\% \\ \hline
        2 & 2 & 66.87\% & -16.27\% & 80.42\% & -6.32\% & 83.85\% & -5.17\% \\ \hline
        1 & 2 & 60.06\% & -24.80\% & 75.29\% & -12.29\% & 79.75\% & -9.80\% \\ \hline
        \midrule
        12 & 1 & 57.40\% & -28.13\% & 73.24\% & -14.68\% & 78.56\% & -11.15\% \\ \hline
        9 & 1 & 57.51\% & -27.99\% & 73.24\% & -14.68\% & 77.87\% & -11.94\% \\ \hline
        6 & 1 & 57.26\% & -28.30\% & 73.52\% & -14.35\% & 78.34\% & -11.40\% \\ \hline
        3 & 1 & 57.04\% & -28.58\% & 73.93\% & -13.87\% & 78.39\% & -11.34\% \\ \hline
        2 & 1 & 56.57\% & -29.17\% & 73.71\% & -14.13\% & 77.98\% & -11.81\% \\ \hline
        1 & 1 & 54.96\% & -31.18\% & 71.88\% & -16.26\% & 76.73\% & -13.22\% \\ \hline
    \end{tabular}}
    \label{tab:asym-nq}
\end{table}
\subsection{Inference Benchmarks}
To evaluate the impact of structural pruning, we benchmark inference speeds of query encoding while varying the number of transformer layers. We perform benchmarking using an Intel Xeon Gold 6238R Processor and a T4 Nvidia GPU. For each model, we evaluate the performance on encoding 6500 queries with a batch size of one and a max context length of 32. For CPU inference, we evaluate the performance of models using the ONNX library \footnote{https://onnx.ai/}, and for GPU inference, we evaluate native Pytorch inference. We repeat each run five times to ensure consistency and report the mean. Summary statistics can be found in table \ref{tab:inference-summary} and full results, including percentile, standard deviation, and confidence intervals, can be found in the appendix \ref{sec:inference-benchmarks}.\\
\begin{table}[!ht]
    \centering
    \tiny
    \begin{tabular}{|l|l|l|l|l|l|}
    \hline
        layers & size & compressed size & method & QPS & Speedup \\ \hline
        12 & 418 & 387 & GPU & 105.852 & 1.00 \\ \hline
        9 & 337 & 212 & GPU & 139.494 & 1.32 \\ \hline
        6 & 256 & 236 & GPU & 172.338 & 1.63 \\ \hline
        3 & 175 & 161 & GPU & 299.45 & 2.83 \\ \hline
        2 & 148 & 136 & GPU & 441.422 & 4.17 \\ \hline
        1 & 121 & 111 & GPU & 660.64 & 6.24 \\ \hline
        12 & 418 & 387 & CPU & 47.278 & 1.00 \\ \hline
        9 & 337 & 212 & CPU & 63.24 & 1.34 \\ \hline
        6 & 256 & 236 & CPU & 90.386 & 1.91 \\ \hline
        3 & 175 & 161 & CPU & 166.012 & 3.51 \\ \hline
        2 & 148 & 136 & CPU & 229.666 & 4.86 \\ \hline
        1 & 121 & 111 & CPU & 378.534 & 8.01 \\ \hline
    \end{tabular}
    \caption{Variation in model throughput according to the serving method and the number of transformer layers. Structural pruning can lead to a 6 and 8-layer performance increase on GPU and CPU and pruning a model to 3 layers allows a CPU to offer better inference performance than the GPU.}
    \label{tab:inference-summary}
\end{table}
\section{KL Alignment of Embeddings}
\begin{table}[!htb]
    \centering
    \caption{Impact of structural pruning with and without KALE on Accuracy at 100 across various datasets. }
    \tiny
    \scalebox{0.8}{
    \begin{tabular}{|l|l|l|l|l|l|l|}
    \hline
        Layers & KALE & NQ & TriviaQA & MSMARCO & SCIFACT & SQUAD \\ \hline
        12 & N/A & 85.84\% & 85.84\% & 88.77\% & 90.70\% & 77.16\% \\ \hline
        9 & N & 79.97\% & 79.97\% & 82.01\% & 71.07\% & 71.38\% \\ \hline
        9 & Y & 84.90\% & 84.90\% & 86.16\% & 84.87\% & 73.54\% \\ \hline
        6 & N & 68.20\% & 68.20\% & 72.68\% & 22.98\% & 59.97\% \\ \hline
        6 & Y & 83.68\% & 83.68\% & 84.68\% & 85.13\% & 69.87\% \\ \hline
        3 & N & 43.88\% & 43.88\% & 11.39\% & 40.80\% & 34.42\% \\ \hline
        3 & Y & 81.14\% & 81.14\% & 82.11\% & 82.57\% & 64.37\% \\ \hline
        2 & N & 46.90\% & 46.90\% & 31.46\% & 42.66\% & 37.01\% \\ \hline
        2 & Y & 81.94\% & 81.94\% & 81.96\% & 82.57\% & 63.72\% \\ \hline
        1 & N & 12.22\% & 12.22\% & 0.00\% & 3.17\% & 11.66\% \\ \hline
        1 & Y & 71.33\% & 71.33\% & 54.36\% & 66.83\% & 51.39\% \\ \hline
    \end{tabular}}
    \label{tab:kale-at-20}
\end{table}
While training asymmetric models can improve latency, it requires novel training regimes and experimentation, and existing workloads need to regenerate their entire index to take advantage of any inference speedups. Generation of the passage index can take longer than model training \cite{Karpukhin2020DensePR}, which makes regenerating a new index and retraining a model to meet changing latency requirements an inefficient experimentation pathway. \\Moreover, coupling asymmetry into training makes generating query encoder variants more difficult, as each encoder requires its own index and document encoder. \\
Motivated by this bottleneck, we introduce \textbf{K}ullback-Leibler \textbf{Al}lingment of \textbf{E}mbeddings (KALE), a simple method of improving bi-encoder latency by aligning the embeddings of compressed models. KALE is applied after model training and leverages large batch sizes to make compression \textbf{computationally inexpensive} and \textbf{independent of training}. A single V100 GPU KALE can produce a compressed query encoder in less than 5 minutes.  \\
First, a bi-encoder model trains with separate query and document encoders. When training is complete, the document encoder, $e_{document}$, is frozen, and using the query encoder, $e_{q}$, a structurally pruned copy, $e_{q'}$, is made. Then, using a sample of queries, the $e_{q'}$ model is fine-tuned to minimize the KL divergence of their query representations as shown in equation \ref{kale-eq:1}. While the KL divergence is a measure of differences in probability distributions it has been applied successfully for representation alignment \cite{Kim2023EmbedDistillAG}. To leverage it, we treat each of the representation vectors as a probability over a set of logits. \\
\begin{equation}\label{kale-eq:1} 
    \displaystyle D_{\text{KL}}(e_{q'} \parallel e_{q} )= \sum _{x\in {\mathcal {X}}}e_{q'}(x)\log \left({\frac {e_{q'}(x)}{e_{q}(x)}}\right).
\end{equation}
We explored the use of various distance functions such as cosine similarity, Manhattan distance, and the KL divergence but found little sensitivity in any metric besides KL divergence. We believe this is due to us freezing the document representations, and as a result, cosine distance allows the query embeddings to \textit{drift} more than probability distribution matching methods. To explore this further, we experiment with tuning the temperature for the KL divergence and add a loss scaling factor but find a temperature of one and a scaling factor of ten to be most optimal. \\
Additionally, we explored using a contrastive loss with random negative and hard negatives mined from the trained encoder but found no positive impact for either method. We leave further exploration of training objective improvement for future work.
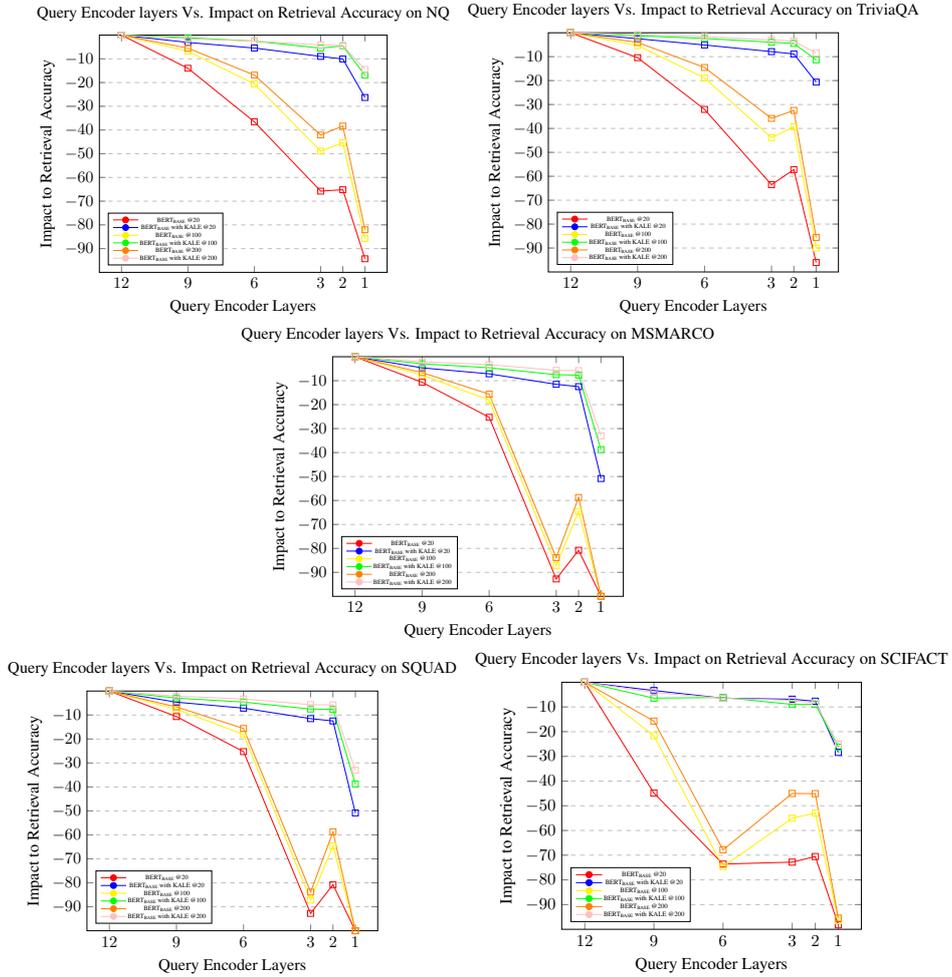
\begin{figure*}
    \centering
    \begin{subfigure}[b]{0.35\textwidth}
   \begin{adjustbox}{width=\linewidth}
      \begin{tikzpicture}
\begin{axis}[
    title={Query Encoder layers Vs. Impact on Retrieval Accuracy on NQ},
    xlabel={Query Encoder Layers},
    ylabel={Impact to Retrieval Accuracy},
    xmin=0, xmax=13,
    x dir=reverse,
    ymin=-100 , ymax=0,
    xtick={1,2,3,6,9,12},
    ytick={-10,-20,-30,-40, -50,-60,-70,-80,-90},
    legend pos=south west,
    ymajorgrids=true,
    grid style=dashed,
    legend style={nodes={scale=0.4, transform shape}}, 
    legend image post style={mark=*}
]
\addplot[
    color=red,
    mark=square,
    ]
    coordinates {
    (12, 0) (9, -13.97) (6,-36.53) (3,-65.77) (2, -65.18) (1, -94.28)
    };

\addplot[
    color=blue,
    mark=square,
    ]
    coordinates {
    (12, 0) (9, -3.08) (6, -5.45) (3,-8.98) (2,-10.06) (1, -26.30)

    };
\addplot[
    color=yellow,
    mark=square,
    ]
    coordinates {
    (12, 0) (9, -6.84) (6,-20.55) (3,-48.88) (2, -45.36) (1, -85.76)
    };

\addplot[
    color=green,
    mark=square,
    ]
    coordinates {
    (12, 0) (9, -1.10) (6, -2.52) (3,-5.48) (2,-4.54) (1, -16.90)

    };
\addplot[
    color=orange,
    mark=square,
    ]
    coordinates {
    (12, 0) (9, -5.51) (6,-16.85) (3,-42.11) (2, -38.32) (1, -82.05)
    };

\addplot[
    color=pink,
    mark=square,
    ]
    coordinates {
    (12, 0) (9, -1.56) (6, -2.53) (3,-4.14) (2,-4.39) (1, -14.44)

    };
\legend{BERT\textsubscript{BASE} @20, BERT\textsubscript{BASE} with KALE @20,BERT\textsubscript{BASE} @100, BERT\textsubscript{BASE} with KALE @100,BERT\textsubscript{BASE} @200, BERT\textsubscript{BASE} with KALE @200}
 \end{axis}
\end{tikzpicture}
 \end{adjustbox} 
\end{subfigure} 
    \begin{subfigure}[b]{0.382\textwidth}
   \begin{adjustbox}{width=\linewidth} 
      \begin{tikzpicture}
        \begin{axis}[
            title={Query Encoder layers Vs. Impact to Retrieval Accuracy on TriviaQA},
            xlabel={Query Encoder Layers},
            ylabel={Impact to Retrieval Accuracy},
            xmin=0, xmax=13,
            x dir=reverse,
            ymin=-100 , ymax=0,
            xtick={1,2,3,6,9,12},
            ytick={-10,-20,-30,-40, -50,-60,-70,-80,-90},
            legend pos=south west,
            ymajorgrids=true,
            grid style=dashed,
            legend style={nodes={scale=0.4, transform shape}}, 
            legend image post style={mark=*}
        ]
        \addplot[
            color=red,
            mark=square,
            ]
            coordinates {
            (12, 0) (9, -10.41) (6,-32.04) (3,-63.5) (2, -57.22) (1, -96.03)
            };
        
        \addplot[
            color=blue,
            mark=square,
            ]
            coordinates {
            (12, 0) (9, -2.48) (6, -5.11) (3,-7.88) (2,-8.86) (1, -20.63)
        
            };
        \addplot[
            color=yellow,
            mark=square,
            ]
            coordinates {
            (12, 0) (9, -5.35) (6,-18.91) (3,-43.84) (2, -39.29) (1, -90.02)
            };
        
        \addplot[
            color=green,
            mark=square,
            ]
            coordinates {
            (12, 0) (9, -1.28) (6,-2.38) (3,-3.95) (2, -4.47) (1, -11.33)
            };
        
        \addplot[
            color=orange,
            mark=square,
            ]
            coordinates {
            (12, 0) (9, -4.04) (6, -14.52) (3,-35.8) (2,-32.45) (1, -85.58)
        
            };
        \addplot[
            color=pink,
            mark=square,
            ]
            coordinates {
            (12, 0) (9, -0.78) (6,-1.59) (3,-2.99) (2, -3.45) (1, -8.54)
            };
        \legend{BERT\textsubscript{BASE} @20, BERT\textsubscript{BASE} with KALE @20,BERT\textsubscript{BASE} @100, BERT\textsubscript{BASE} with KALE @100,BERT\textsubscript{BASE} @200, BERT\textsubscript{BASE} with KALE @200}
         \end{axis}
        \end{tikzpicture}
         \end{adjustbox}       
    \end{subfigure}
    \\
    \begin{subfigure}[b]{0.4\textwidth}
       \begin{adjustbox}{width=\linewidth} 
          \begin{tikzpicture}
            \begin{axis}[
                title={Query Encoder layers Vs. Impact to Retrieval Accuracy on MSMARCO},
                xlabel={Query Encoder Layers},
                ylabel={Impact to Retrieval Accuracy},
                xmin=0, xmax=13,
                x dir=reverse,
                ymin=-100 , ymax=0,
                xtick={1,2,3,6,9,12},
                ytick={-10,-20,-30,-40, -50,-60,-70,-80,-90},
                legend pos=south west,
                ymajorgrids=true,
                grid style=dashed,
                legend style={nodes={scale=0.4, transform shape}}, 
                legend image post style={mark=*}
            ]
            \addplot[
                color=red,
                mark=square,
                ]
                coordinates {
                (12, 0) (9, -10.65) (6,-25.27) (3,-92.76) (2, -80.77) (1, -100)
                };
            
            \addplot[
                color=blue,
                mark=square,
                ]
                coordinates {
                (12, 0) (9, -4.64) (6, -7.14) (3,-11.51) (2,-12.48) (1, -50.84)
            
                };
            \addplot[
                color=yellow,
                mark=square,
                ]
                coordinates {
                (12, 0) (9, -7.62) (6,-18.12) (3,-87.17) (2, -64.56) (1, -100)
                };
            
            \addplot[
                color=green,
                mark=square,
                ]
                coordinates {
                (12, 0) (9, -2.94) (6, -4.6) (3,-7.5) (2,-7.67) (1, -38.77)
            
                };
            \addplot[
                color=orange,
                mark=square,
                ]
                coordinates {
                (12, 0) (9, -6.63) (6,-15.6) (3,-83.85) (2, -58.75) (1, -100)
                };
            
            \addplot[
                color=pink,
                mark=square,
                ]
                coordinates {
                (12, 0) (9, -2.12) (6, -3.3) (3,-5.68) (2,-5.79) (1, -33.05)
            
                };
            \legend{BERT\textsubscript{BASE} @20, BERT\textsubscript{BASE} with KALE @20,BERT\textsubscript{BASE} @100, BERT\textsubscript{BASE} with KALE @100,BERT\textsubscript{BASE} @200, BERT\textsubscript{BASE} with KALE @200}
             \end{axis}
            \end{tikzpicture}
        \end{adjustbox}       
    \end{subfigure} \\ 
    \begin{subfigure}[b]{0.38\textwidth}
       \begin{adjustbox}{width=\linewidth} 
          \begin{tikzpicture}
            \begin{axis}[
                title={Query Encoder layers Vs. Impact on Retrieval Accuracy on SQUAD},
                xlabel={Query Encoder Layers},
                ylabel={Impact to Retrieval Accuracy},
                xmin=0, xmax=13,
                x dir=reverse,
                ymin=-100 , ymax=0,
                xtick={1,2,3,6,9,12},
                ytick={-10,-20,-30,-40, -50,-60,-70,-80,-90},
                legend pos=south west,
                ymajorgrids=true,
                grid style=dashed,
                legend style={nodes={scale=0.4, transform shape}}, 
                legend image post style={mark=*}
            ]
            \addplot[
                color=red,
                mark=square,
                ]
                coordinates {
                (12, 0) (9, -10.65) (6,-25.27) (3,-92.76) (2, -80.77) (1, -100)
                };
            
            \addplot[
                color=blue,
                mark=square,
                ]
                coordinates {
                (12, 0) (9, -4.64) (6, -7.14) (3,-11.51) (2,-12.48) (1, -50.84)
            
                };
            \addplot[
                color=yellow,
                mark=square,
                ]
                coordinates {
                (12, 0) (9, -7.62) (6,-18.12) (3,-87.17) (2, -64.56) (1, -100)
                };
            
            \addplot[
                color=green,
                mark=square,
                ]
                coordinates {
                (12, 0) (9, -2.94) (6, -4.6) (3,-7.5) (2,-7.67) (1, -38.77)
            
                };
            \addplot[
                color=orange,
                mark=square,
                ]
                coordinates {
                (12, 0) (9, -6.63) (6,-15.6) (3,-83.85) (2, -58.75) (1, -100)
                };
            
            \addplot[
                color=pink,
                mark=square,
                ]
                coordinates {
                (12, 0) (9, -2.12) (6, -3.3) (3,-5.68) (2,-5.79) (1, -33.05)
            
                };
            \legend{BERT\textsubscript{BASE} @20, BERT\textsubscript{BASE} with KALE @20,BERT\textsubscript{BASE} @100, BERT\textsubscript{BASE} with KALE @100,BERT\textsubscript{BASE} @200, BERT\textsubscript{BASE} with KALE @200}
             \end{axis}
            \end{tikzpicture}
        \end{adjustbox}
    \end{subfigure} 
    \begin{subfigure}[b]{0.4\textwidth}
       \begin{adjustbox}{width=\linewidth} 
          \begin{tikzpicture}
            \begin{axis}[
                title={Query Encoder layers Vs. Impact on Retrieval Accuracy on SCIFACT},
                xlabel={Query Encoder Layers},
                ylabel={Impact to Retrieval Accuracy},
                xmin=0, xmax=13,
                x dir=reverse,
                ymin=-100 , ymax=0,
                xtick={1,2,3,6,9,12},
                ytick={-10,-20,-30,-40, -50,-60,-70,-80,-90},
                legend pos=south west,
                ymajorgrids=true,
                grid style=dashed,
                legend style={nodes={scale=0.4, transform shape}}, 
                legend image post style={mark=*}
            ]
            \addplot[
                color=red,
                mark=square,
                ]
                coordinates {
                (12, 0) (9, -44.85) (6,-73.60) (3,-72.81) (2, -70.55) (1, -98.18)
                };
            
            \addplot[
                color=blue,
                mark=square,
                ]
                coordinates {
                (12, 0) (9, -3.34) (6,-6.45) (3,-6.86) (2, -7.66) (1, -28.38)
                };
            \addplot[
                color=yellow,
                mark=square,
                ]
                coordinates {
                (12, 0) (9, -21.64) (6,-74.66) (3,-55.02) (2, -52.95) (1, -96.50)
                };
            
            \addplot[
                color=green,
                mark=square,
                ]
                coordinates {
                (12, 0) (9, -6.43) (6,-6.14) (3,-8.96) (2, -8.96) (1, -26.32)
                };
            \addplot[
                color=orange,
                mark=square,
                ]
                coordinates {
                (12, 0) (9, -15.72) (6,-67.83) (3,-45.01) (2, -45.09) (1, -95.49)
                };
            
            \addplot[
                color=pink,
                mark=square,
                ]
                coordinates {
                (12, 0) (9, -4.13) (6,-6.47) (3,-7.33) (2, -8.39) (1, -25.11)
                };
            \legend{BERT\textsubscript{BASE} @20, BERT\textsubscript{BASE} with KALE @20,BERT\textsubscript{BASE} @100, BERT\textsubscript{BASE} with KALE @100,BERT\textsubscript{BASE} @200, BERT\textsubscript{BASE} with KALE @200}
             \end{axis}
            \end{tikzpicture}
        \end{adjustbox}
    \end{subfigure}
    \caption{Impact of structural pruning with and without KALE on the NQ, MSMARCO, TriviaQA, SciFACT, and SQuAD Passage Retrieval dataset with the recall set sizes of 20,100, and 200. Across datasets, we see a consistent trend where KALE is effective but most effective when the network is heavily pruned and recall set sizes are small. When the model is pruned to 2 or 1 layer with a recall set size of 20, the difference between using KALE or not can be up to 10 times the loss in recall accuracy}
    \label{fig:kale-not}
\end{figure*}
\subsection{Experimental Results}
We evaluate the effectiveness of KALE by taking uncompressed BERT\textsubscript{BASE} models and pruning them with and without KALE on a variety of well-established passage retrieval benchmarks. First, models are trained, and indexes are generated using un-optimized BERT\textsubscript{BASE} models. Next, the document encoders are frozen, and the query encoders are structurally pruned to have 9,6,3,2 or 1 transformer layer. Finally, query encoders are aligned using KALE, and we compare the performance of compressed models by comparing the impact on retrieval accuracy at 20,100, and 200. \\
To aid reproducibility, each model is trained using the Tevatron \cite{Gao2022TevatronAE} \footnote{https://github.com/texttron/tevatron} library, which makes use of hugginface's transformers to provide a simple interface for exploring neural ranking models. Our experiments focus on the plain BERT\textsubscript{BASE}-uncased 12-layer transformer model. While never more capable models exist, the unaltered BERT model is widely used in production workloads, which our experiments seek to emulate. \\
Our work aims not to produce the highest possible retrieval accuracy for a dense encoder. Instead, our goal is to find the role of asymmetry in bi-encoder models. As a result, we leverage the well-established parameters in all of our experiments without using an advanced methodology like contrastive or curriculum learning. \\
There are fewer parameters for using KALE, and we deliberately do not optimize on anything but the loss between $e_{q}$ and $e_{q'}$. In general, higher degrees of pruning require longer training with smaller batches. \\ 
\textbf{Datasets} We use a wide variety of standard dense retrieval benchmarks, including MSMARCO V1.1 \footnote{https://huggingface.co/datasets/Tevatron/msmarco-passage} \cite{Campos2016MSMA}, NQ Passage Ranking \footnote{https://huggingface.co/datasets/Tevatron/wikipedia-nq} \cite{Kwiatkowski2019NaturalQA}, SciFact Passage Ranking \footnote{https://huggingface.co/datasets/Tevatron/scifact} \cite{Wadden2020FactOF}, TriviaQA passage Ranking \footnote{https://huggingface.co/datasets/Tevatron/wikipedia-trivia} \cite{Joshi2017TriviaQAAL}, and SQUAD Passage Ranking \footnote{https://huggingface.co/datasets/Tevatron/wikipedia-squad} \cite{Rajpurkar2016SQuAD10}. \\
For each dataset, we evaluate performance by measuring the recall accuracy with retrieval depths of 20,100, and 200. Additionally, for the MSMARCO dataset, we also report MRR@10; for Scifact, we also report NDCG @10 and RR@10. \\
\textbf{Computational Experiments}
Our experimentation on fine-tuning our compressed models uses a 16 GB V100 GPU. Experiments in bi-encoder model training leverage 1 V100 for the MSMARCO and 4 for each other experiment. Due to the vast number of models and datasets we train on, each experiment happens with the same fixed seed.  

\subsection{Evaluating KALE}
We compare the performance of using KALE for post-training compression in figure \ref{fig:kale-not} across the five datasets and see a fairly consistent trend. When the recall set is small and the query encoders are pruned to a high degree, the impact of KALE is most visible, often driving over 50 improvements in retrieval accuracy. Additionally, using KALE allows the models to have a steady and gradual drop in recall accuracy relative to speedup instead of the sharp drop shown by the regular usage of structural pruning.  Without KALE, post-training compression causes a 20-50\% loss in retrieval accuracy. With the use of KALE, these losses are cut to 1-10\%. In practice, this allows using one or 2-layer encoder models running with CPU-based inference with minor impacts on accuracy. \\
\begin{table}[!ht]
    \centering
    \tiny
    \scalebox{0.68}{
    \begin{tabular}{|l|l|l|l|l|l|l|l|}
    \hline
        Model &Layers& KALE & MSMARCO & NQ & TriviaQA & SQUAD & SCIFACTS \\ \hline
        BERT\textsubscript{BASE} & 12 & N & 88.77\% & 85.84\% & 85.03\% & 77.16\% & 90.70\% \\ \hline
        \midrule
        BERT\textsubscript{BASE} & 6 & Y & 84.68\% & 83.68\% & 83.01\% & 69.87\% & 85.13\% \\ \hline
        $6_{kd}-6_{kd}$ & 6 & N & 88.19\% & 85.15\% & 84.96\% & 71.94\% & 91.23\% \\ \hline
        $6_{db}-6_{db}$ & 6 & N & 88.35\% & 84.74\% & 84.83\% & 71.69\% & 89.37\% \\ \hline
        $6_{kd}-3_{kd}$ & 6 & N & 86.50\% & 85.37\% & 84.04\% & 70.89\% & 89.20\% \\ \hline
        \midrule
        BERT\textsubscript{BASE} & 3 & Y & 82.11\% & 81.14\% & 81.67\% & 64.37\% & 82.57\% \\ \hline
        $3_{kd}-3_{kd}$ & 3 & N & 86.13\% & 83.66\% & 84.11\% & 71.98\% & 89.40\% \\ \hline
        $3_{kd}-6_{kd}$ & 3 & N & 84.79\% & 85.76\% & 83.91\% & 67.85\% & 88.63\% \\ \hline
        $6_{kd}-3_{kd}$ & 3 & Y & 82.95\% & 83.43\% & 82.33\% & 63.77\% & 90.37\% \\ \hline
        $6_{kd}-6_{kd}$ & 3 & Y & 86.75\% & 80.78\% & 83.48\% & 64.14\% & 91.70\% \\ \hline
        \midrule
        BERT\textsubscript{BASE} & 2 & Y & 81.96\% & 81.94\% & 81.23\% & 67.00\% & 82.57\% \\ \hline
        $3_{kd}-3_{kd}$ & 2 & Y & 84.23\% & 82.71\% & 83.02\% & 67.02\% & 91.33\% \\ \hline
        $3_{kd}-6_{kd}$ & 2 & Y & 85.57\% & 84.27\% & 82.90\% & 62.75\% & 88.37\% \\ \hline
        $6_{kd}-3_{kd}$ & 2 & Y & 83.24\% & 83.02\% & 82.13\% & 62.52\% & 89.93\% \\ \hline
        $6_{kd}-6_{kd}$ & 2 & Y & 85.77\% & 80.39\% & 83.32\% & 52.74\% & 91.93\% \\ \hline 
        \midrule
        BERT\textsubscript{BASE} & 1 & Y & 48.05\% & 71.33\% & 75.40\% & 51.39\% & 66.83\% \\ \hline
        $3_{kd}-3_{kd}$ & 1 & Y & 66.69\% & 77.17\% & 80.82\% & 55.62\% & 76.03\% \\ \hline
        $3_{kd}-6_{kd}$ & 1 & Y & 72.13\% & 79.81\% & 80.23\% & 52.26\% & 78.67\% \\ \hline
        $6_{kd}-3_{kd}$ & 1 & Y & 71.26\% & 76.57\% & 78.65\% & 50.88\% & 77.07\% \\ \hline
        $6_{kd}-6_{kd}$ & 1 & Y & 70.70\% & 74.71\% & 80.31\% & 52.74\% & 77.89\% \\ \hline
    \end{tabular}}
    \caption{Impact of model asymmetry and use of KALE for structural pruning on the Retrieval at 100 accuracies across various datasets. Layers refer to the number of transformer encoder layers in the query encoder.  }
    \label{tab:kale+asym-100}
\end{table}
We also notice a surprising performance improvement between 3 and 2-layer query encoders with and without KALE. We believe this shows the phenomena studied elsewhere: the first and last layers do most of the work \cite{Oh2022DontJA}. 
\subsection{Aiding Asymmetry with KALE}
Seeking to optimize compression further, we combine KALE with asymmetrical finetuning and evaluate the results similarly to our earlier experiments. Results on the impact of KALE and asymmetry on the five datasets on the recall accuracy at 100 can be found in table \ref{tab:kale+asym-100} where $3_{kd}-6_{kd}$ denotes a three-layer query encoder and six-layer document encoder, $3_{kd}-3_{kd}$ denotes dual three layer encoders. Full results and metrics for each task can be found in the appendix section \ref{sec:kale-asym-full}. \\
\begin{figure}[!htb]
\begin{tikzpicture}
\scalebox{0.85}{
\begin{axis}[
    title={Inference Speed (GPU) Vs.Retrieval Accuracy @100   },
    xlabel={Queries Per Second},
    ylabel={Retrieval Accuracy},
    xmin=90, xmax=700,
    ymin=50 , ymax=95,
    xtick={100, 200, 300,400,500,600},
    ytick={60,70,80,90},
    legend pos=north east,
    ymajorgrids=true,
    grid style=dashed,
    legend style={nodes={scale=0.4, transform shape}}, 
    legend image post style={mark=*}
]
\addplot[
    color=black,
    mark=square,
    ]
    coordinates {
    (105, 88.77) (172,88.35) (300,86.13) (441, 85.77) (660, 72.13)
    };
\addplot[
    color=blue,
    mark=square,
    ]
    coordinates {
    (105, 85.84) (172,85.15) (300,85.76) (441, 84.27) (660, 79.81)
    };
\addplot[
    color=red,
    mark=square,
    ]
    coordinates {
    (105, 85.03) (172,84.96) (300,84.11) (441, 83.32) (660, 80.82)
    };
\addplot[
    color=green,
    mark=square,
    ]
    coordinates {
    (105,77.16) (172,71.94) (300,71.98) (441,67.02) (660, 55.62)
    };
\addplot[
    color=brown,
    mark=square,
    ]
    coordinates {
    (105, 90.7) (172,91.23) (300,91.7) (441, 91.93) (660, 78.67)
    };

\legend{MSMARCO, NQ, TriviaQA, SQUAD, SCIfacts }
 \end{axis}}

\end{tikzpicture}
    \centering
    \caption{The impact on retrieval accuracy of the best combinations of asymmetrical training and KALE across the NQ, MSMARCO, TriviaQA, SQUAD, and SCIfacts retrieval datasets}
    \label{fig:speed-vs-acc}
\end{figure}
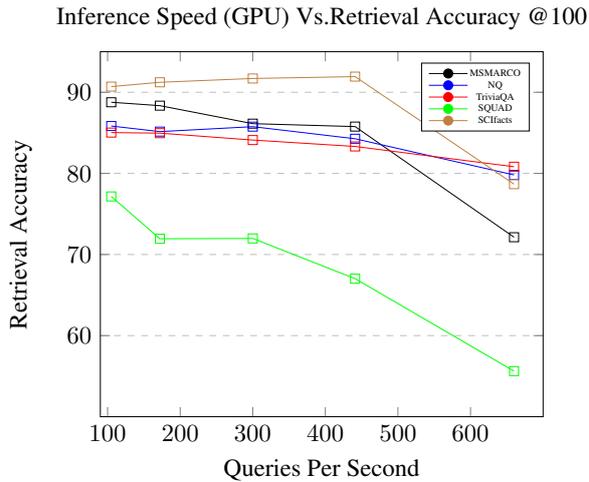
First, it is immediately observable that post-training compression via KALE performs worse than models natively designed for that size. We believe this is due to the convergence of the KALE models to have \textit{some distance} from the uncompressed model because of dropout. We experimented with not using dropout in KALE, but model performance quickly suffered. \\
Looking at the best retrieval accuracy vs. the model speedups shown in figure \ref{fig:speed-vs-acc}, we can see a substantial variation in the impact of compression across datasets. In tasks like SCIfacts, it is possible to get over 4x speedup while improving accuracy, while on tasks like SQuAD, even minor speedups lead to major losses in accuracy. We believe this variation is driven by the relative difficulty of each dataset, where easier tasks are more compressible than harder tasks. \\
We believe these variations in results highlight the utility of post-training compression methods like KALE. Given the task variability in the impact of compression, iteration speed and cost are essential to effectively tuning model inference speed and accuracy. 
\section{Limitations}
While our work makes a broad study on how to improve model efficiency our scope is limited. Our work is limited to the usage of BERT-base and it is not clear how our compression approaches scale to more varied architectures like the sequence-to-sequence models used by DocT5 \cite{lee-etal-2022-docmt5} or more optimized models like RoBERTa \cite{Liu2019RoBERTaAR} or compressed models like MiniLM \cite{wang2020minilm}.
\section{Conclusion and Future Work}
In this work, we have demonstrated how the use of asymmetry between the query and document encoders in bi-encoder models can be leveraged for improved inference efficiencies across CPUs and GPUs. Using our post-training compression framework, KALE, we can compress models up to 6x with little loss in accuracy. Compressing models without regenerating the document index or the document encoder makes it practical to have many query encoders tailored to each use case's latency needs. \\
In the future, we wish to study how asymmetry in retrieval can be implemented with models which are widely different and may have different hidden sizes, such as using MiniLM for the query model and RoBERTA-Large for the document model. 
\bibliographystyle{acl_natbib}
\bibliography{anthology,custom}
\appendix
\subsection{Asymmetrical Dense Retrieval}
the impact of structural pruning with asymmetrical dense retrieval can be found in table \ref{tab:asym-nq-kd}. Similar to other works studying the use of knowledge distillation found \cite{Sanh2020MovementPA}, the use of distillation improves performance by a non-negligible level.
\begin{table}[!htb]
    \centering
    \caption{Impact of Structural pruning with knowledge distilled variants before fine-tuning on Retrieval Accuracy on NQ passage retrieval dataset}
    \scalebox{0.5}{
    \begin{tabular}{|l|l|l|l|l|l|l|l|}
    \hline
        $layers_{q}$ & $layers_{d}$ & Top 20 & Impact & Top 100 & Impact & Top 200 & Impact \\ \hline
        12 & 12 & 79.86\% & 0.00\% & 85.84\% & 0.00\% & 88.42\% & 0.00\% \\ \hline
        \midrule
        $6_{distilbert}$ & $6_{distilbert}$ & 73.88\% & -7.49\% & 84.74\% & -1.29\% & 87.26\% & -1.31\% \\ \hline
        $6_{KD}$ & 12 & 73.99\% & -7.35\% & 84.32\% & -1.77\% & 86.65\% & -2.00\% \\ \hline
        $6_{KD}$ & 9 & 71.63\% & -10.30\% & 83.16\% & -3.12\% & 85.82\% & -2.94\% \\ \hline
        $6_{KD}$ & 6 & 71.00\% & -11.10\% & 82.35\% & -4.06\% & 85.48\% & -3.32\% \\ \hline
        $6_{KD}$ & 3 & 68.42\% & -14.32\% & 80.94\% & -5.71\% & 84.24\% & -4.73\% \\ \hline
        $6_{KD}$ & 2 & 68.39\% & -14.36\% & 80.58\% & -6.13\% & 84.02\% & -4.98\% \\ \hline
        $6_{KD}$ & 1 & 56.62\% & -29.10\% & 72.24\% & -15.84\% & 77.81\% & -12.00\% \\ \hline
        \midrule
        $3_{KD}$ & 12 & 71.72\% & -10.20\% & 83.21\% & -3.06\% & 85.90\% & -2.85\% \\ \hline
        $3_{KD}$ & 9 & 68.95\% & -13.66\% & 81.75\% & -4.77\% & 84.79\% & -4.10\% \\ \hline
        $3_{KD}$ & 6 & 68.09\% & -14.74\% & 81.52\% & -5.03\% & 84.76\% & -4.13\% \\ \hline
        $3_{KD}$ & 3 & 65.84\% & -17.55\% & 79.58\% & -7.29\% & 83.41\% & -5.67\% \\ \hline
        $3_{KD}$ & 2 & 66.81\% & -16.34\% & 79.50\% & -7.38\% & 82.71\% & -6.45\% \\ \hline
        $3_{KD}$ & 1 & 54.46\% & -31.81\% & 71.44\% & -16.77\% & 76.59\% & -13.38\% \\ \hline
        \midrule
        12 & $6_{KD}$ & 78.78\% & -1.35\% & 85.84\% & 0.01\% & 87.45\% & -1.10\% \\ \hline
        9 & $6_{KD}$ & 77.26\% & -3.26\% & 85.18\% & -0.77\% & 87.34\% & -1.22\% \\ \hline
        6 & $6_{KD}$ & 76.45\% & -4.26\% & 84.96\% & -1.03\% & 87.06\% & -1.53\% \\ \hline
        $6_{KD}$ & $6_{KD}$ & 75.04\% & -6.03\% & 85.15\% & -0.80\% & 87.45\% & -1.10\% \\ \hline
        3 & $6_{KD}$ & 74.49\% & -6.73\% & 84.24\% & -1.87\% & 86.54\% & -2.13\% \\ \hline
        $3_{KD}$ & $6_{KD}$ & 77.01\% & -3.57\% & 85.76\% & -0.09\% & 87.42\% & -1.13\% \\ \hline
        2 & $6_{KD}$ & 74.43\% & -6.80\% & 83.68\% & -2.51\% & 86.32\% & -2.38\% \\ \hline
        1 & $6_{KD}$ & 68.09\% & -14.74\% & 79.22\% & -7.71\% & 83.19\% & -5.92\% \\ \hline
        \midrule
        12 & $3_{KD}$ & 76.45\% & -4.26\% & 84.49\% & -1.58\% & 86.70\% & -1.94\% \\ \hline
        9 & $3_{KD}$ & 76.12\% & -4.68\% & 84.29\% & -1.80\% & 86.26\% & -2.44\% \\ \hline
        6 & $3_{KD}$ & 75.15\% & -5.89\% & 83.43\% & -2.80\% & 86.45\% & -2.22\% \\ \hline
        $6_{KD}$ & $3_{KD}$ & 77.40\% & -3.09\% & 85.37\% & -0.54\% & 87.48\% & -1.06\% \\ \hline
        $3_{KD}$ & $3_{KD}$ & 73.32\% & -8.18\% & 83.43\% & -2.80\% & 86.20\% & -2.51\% \\ \hline
        3 & $3_{KD}$ & 71.88\% & -9.99\% & 83.66\% & -2.54\% & 86.37\% & -2.32\% \\ \hline
        2 & $3_{KD}$ & 72.22\% & -9.56\% & 81.93\% & -4.55\% & 85.08\% & -3.77\% \\ \hline
        1 & $3_{KD}$ & 67.31\% & -15.71\% & 79.25\% & -7.67\% & 82.77\% & -6.39\% \\ \hline
    \end{tabular}}
    \label{tab:asym-nq-kd}
\end{table}
\subsection{Dense Retrieval and KALE Hyperparameters}
Our experiments focus on minimal hyperparameter optimization. For training of the dense retrievers, we use the datasets described in \ref{tab:hyperparams-kale} where the shorter training lengths and smaller batch sizes correspond to MSMARCO while the other datasets leverage the longer and larger training. For the use of KALE we perform task-specific grid search using the parameters described by \ref{tab:hyperparams-transfer}.
\begin{table}
      \centering
        {\small 
            \begin{tabular}{l|c}
            \toprule
            Parameter & Possible Values \\
            \midrule
            Training Length& 3,40 Epochs \\
            Initial learning rate & 1e-5, 5e-5, 5e-6\\
            Learning rate schedule &  Linear \\
            \greyrule
                Batch size & 8,128, \\
            \greyrule
            \greyrule
                Negative Passages & 1,8 \\
            \greyrule
            \bottomrule
            \end{tabular}
        }
    \caption{Hyperparmaters used to train bi-encoder models for retrieval    \label{tab:hyperparams-kale}}
\end{table}
\begin{table}
      \centering
        {\small 
            \begin{tabular}{l|c}
            \toprule
            Parameter & Possible Values\\
            \midrule
            Training Length & 1,10,100 Epochs  \\
            Initial learning rate & 5e-5, 5e-4, 5e-6\\
            Learning rate schedule &  constant \\
            \greyrule
                Batch size & 4,64,256 \\
            \greyrule
            \greyrule
                Loss Temperature & 1, 10 \\
            \greyrule
            \bottomrule
            \end{tabular}
        }
    \caption{Hyperparmaters used by KALE for aligning the embeddings of a pruned model with its uncompressed target.\label{tab:hyperparams-transfer}}
\end{table}
\subsection{KALE}
\label{sec:KALE-full}
As shown in table \ref{tab:kale-nq}, we explore the impact of KALE for the NQ dataset, in table \ref{tab:kale-trivia}, we explore the impact on TriviaQA, in table \ref{tab:kale-msmarco}, we evaluate the MSARCO passage retrieval, in table \ref{tab:kale-scifacts} we explore Scifacts, and in table \ref{tab:kale-squad} we explore SQUAD. The impact of pruning and KALE is fairly consistent across datasets, but there are larger losses on some smaller datasets, such as SCIfacts and SQUAD. 
\begin{table*}[!ht]
    \centering
    \begin{tabular}{|l|l|l|l|l|l|l|l|}
    \hline
        Layers & KALE & Top 20 & Impact & Top 100 & Impact & Top 200 & Impact \\ \hline
        12 & N/A & 79.86\% & 0.00\% & 85.84\% & 0.00\% & 88.42\% & 0.00\% \\ \hline
        9 & N & 68.70\% & -13.97\% & 79.97\% & -6.84\% & 83.55\% & -5.51\% \\ \hline
        9 & Y & 77.40\% & -3.08\% & 84.90\% & -1.10\% & 87.04\% & -1.56\% \\ \hline
        6 & N & 50.69\% & -36.53\% & 68.20\% & -20.55\% & 73.52\% & -16.85\% \\ \hline
        6 & Y & 75.51\% & -5.45\% & 83.68\% & -2.52\% & 86.18\% & -2.53\% \\ \hline
        3 & N & 27.34\% & -65.77\% & 43.88\% & -48.88\% & 51.19\% & -42.11\% \\ \hline
        3 & Y & 72.69\% & -8.98\% & 81.14\% & -5.48\% & 84.76\% & -4.14\% \\ \hline
        2 & N & 27.81\% & -65.18\% & 46.90\% & -45.36\% & 54.54\% & -38.32\% \\ \hline
        2 & Y & 71.83\% & -10.06\% & 81.94\% & -4.54\% & 84.54\% & -4.39\% \\ \hline
        1 & N & 4.57\% & -94.28\% & 12.22\% & -85.76\% & 15.87\% & -82.05\% \\ \hline
        1 & Y & 58.86\% & -26.30\% & 71.33\% & -16.90\% & 75.65\% & -14.44\% \\ \hline
    \end{tabular}
    \caption{Impact of structural pruning with and without KALE on the NQ retrieval dataset}
    \label{tab:kale-nq}
\end{table*}

\begin{table*}[!ht]
    \centering
    \begin{tabular}{|l|l|l|l|l|l|l|l|}
    \hline
        Layers & KALE & Top 20 & Impact & Top 100 & Impact & Top 200 & Impact \\ \hline
        12 & N/A & 79.43\% & 0.00\% & 85.84\% & 0.00\% & 86.63\% & 0.00\% \\ \hline
        9 & N & 71.16\% & -10.41\% & 79.97\% & -5.35\% & 83.13\% & -4.04\% \\ \hline
        9 & Y & 77.46\% & -2.48\% & 84.90\% & -1.28\% & 85.95\% & -0.78\% \\ \hline
        6 & N & 53.98\% & -32.04\% & 68.20\% & -18.91\% & 74.05\% & -14.52\% \\ \hline
        6 & Y & 75.37\% & -5.11\% & 83.68\% & -2.38\% & 85.25\% & -1.59\% \\ \hline
        3 & N & 28.99\% & -63.50\% & 43.88\% & -43.84\% & 55.62\% & -35.80\% \\ \hline
        3 & Y & 73.17\% & -7.88\% & 81.14\% & -3.95\% & 84.04\% & -2.99\% \\ \hline
        2 & N & 33.98\% & -57.22\% & 46.90\% & -39.29\% & 58.52\% & -32.45\% \\ \hline
        2 & Y & 72.39\% & -8.86\% & 81.94\% & -4.47\% & 83.64\% & -3.45\% \\ \hline
        1 & N & 3.15\% & -96.03\% & 12.22\% & -90.02\% & 12.49\% & -85.58\% \\ \hline
        1 & Y & 63.04\% & -20.63\% & 71.33\% & -11.33\% & 79.23\% & -8.54\% \\ \hline
    \end{tabular}
    \caption{Impact of structural pruning with and without KALE on the TriviaQA retrieval dataset}
    \label{tab:kale-trivia}
\end{table*}

\begin{table*}[!ht]
    \centering
    \scalebox{0.8}{
    \begin{tabular}{|l|l|l|l|l|l|l|l|l|l|}
    \hline
        Layers & KALE & MRR@10 & Impact & Top 20 & Impact & Top 100 & Impact & Top 200 & Impact \\ \hline
        12 & N/A & 32.47\% & 0.00\% & 70.47\% & 0.00\% & 88.77\% & 0.00\% & 93.84\% & 0.00\% \\ \hline
        9 & N & 27.68\% & -14.74\% & 62.97\% & -10.65\% & 82.01\% & -7.62\% & 87.62\% & -6.63\% \\ \hline
        9 & Y & 30.38\% & -6.43\% & 67.21\% & -4.64\% & 86.16\% & -2.94\% & 91.85\% & -2.12\% \\ \hline
        6 & N & 20.86\% & -35.75\% & 52.66\% & -25.27\% & 72.68\% & -18.12\% & 79.20\% & -15.60\% \\ \hline
        6 & Y & 28.71\% & -11.57\% & 65.44\% & -7.14\% & 84.68\% & -4.60\% & 90.74\% & -3.30\% \\ \hline
        3 & N & 1.49\% & -95.42\% & 5.10\% & -92.76\% & 11.39\% & -87.17\% & 15.16\% & -83.85\% \\ \hline
        3 & Y & 26.56\% & -18.19\% & 62.36\% & -11.51\% & 82.11\% & -7.50\% & 88.51\% & -5.68\% \\ \hline
        2 & N & 3.48\% & -89.28\% & 13.55\% & -80.77\% & 31.46\% & -64.56\% & 38.71\% & -58.75\% \\ \hline
        2 & Y & 26.10\% & -19.61\% & 61.68\% & -12.48\% & 81.96\% & -7.67\% & 88.41\% & -5.79\% \\ \hline
        1 & N & 0.00\% & -100.00\% & 0.00\% & -100.00\% & 0.00\% & -100.00\% & 0.00\% & -100.00\% \\ \hline
        1 & Y & 13.16\% & -59.47\% & 34.64\% & -50.84\% & 54.36\% & -38.77\% & 62.82\% & -33.05\% \\ \hline
    \end{tabular}}
    \caption{Impact of structural pruning with and without KALE on the MSMARCO retrieval dataset}
    \label{tab:kale-msmarco}
\end{table*}

\begin{table*}[!ht]
    \centering
    \scalebox{0.60}{
    \begin{tabular}{|l|l|l|l|l|l|l|l|l|l|l|l|l|l|}
    \hline
        Layers & KALE & RR@10 & Impact & recall 10 & Impact & NDCG@10 & Impact & Top 20 & Impact & Top 100 & Impact & Top 200 & Impact \\ \hline
        12 & N/A & 59.11\% & 0.00\% & 78.71\% & 0.00\% & 62.55\% & 0.00\% & 82.38\% & 0.00\% & 90.70\% & 0.00\% & 93.77\% & 0.00\% \\ \hline
        9 & N & 25.30\% & -57.20\% & 39.66\% & -49.61\% & 27.46\% & -56.10\% & 45.43\% & -44.85\% & 71.07\% & -21.64\% & 79.03\% & -15.72\% \\ \hline
        9 & Y & 59.76\% & 1.10\% & 74.86\% & -4.89\% & 62.26\% & -0.46\% & 79.63\% & -3.34\% & 84.87\% & -6.43\% & 89.90\% & -4.13\% \\ \hline
        6 & N & 8.67\% & -85.33\% & 15.06\% & -80.87\% & 9.16\% & -85.36\% & 21.75\% & -73.60\% & 22.98\% & -74.66\% & 30.17\$ &	-67.83\% \\ \hline
        6 & Y & 54.99\% & -6.97\% & 72.53\% & -7.85\% & 58.22\% & -6.92\% & 77.07\% & -6.45\% & 85.13\% & -6.14\% & 87.70\% & -6.47\% \\ \hline
        3 & N & 9.00\% & -84.77\% & 16.00\% & -79.67\% & 9.72\% & -84.46\% & 22.40\% & -72.81\% & 40.80\% & -55.02\% & 51.56\% & -45.01\% \\ \hline
        3 & Y & 55.18\% & -6.65\% & 77.22\% & -1.89\% & 58.30\% & -6.79\% & 76.73\% & -6.86\% & 82.57\% & -8.96\% & 86.90\% & -7.33\% \\ \hline
        2 & N & 9.65\% & -83.67\% & 16.93\% & -78.49\% & 10.39\% & -83.39\% & 24.26\% & -70.55\% & 42.66\% & -52.97\% & 51.49\% & -45.09\% \\ \hline
        2 & Y & 54.45\% & -7.88\% & 71.72\% & -8.88\% & 57.71\% & -7.74\% & 76.07\% & -7.66\% & 82.57\% & -8.96\% & 85.90\% & -8.39\% \\ \hline
        1 & N & 0.30\% & -99.49\% & 13.30\% & -83.10\% & 0.49\% & -99.22\% & 1.50\% & -98.18\% & 3.17\% & -96.50\% & 4.23\% & -95.49\% \\ \hline
        1 & Y & 40.52\% & -31.45\% & 55.25\% & -29.81\% & 43.23\% & -30.89\% & 59.00\% & -28.38\% & 66.83\% & -26.32\% & 70.22\% & -25.11\% \\ \hline
    \end{tabular}}
    \caption{Impact of structural pruning with and without KALE on the SCIFACTS retrieval dataset}
    \label{tab:kale-scifacts}
\end{table*}

\begin{table*}[!ht]
    \centering
    \begin{tabular}{|l|l|l|l|l|l|l|l|}
    \hline
        Layers & KALE & Top 20 & Impact & Top 100 & Impact & Top 200 & Impact \\ \hline
        12 & N/A & 63.82\% & 0.00\% & 77.16\% & 0.00\% & 81.06\% & 0.00\% \\ \hline
        9 & N & 56.16\% & -12.00\% & 71.38\% & -7.49\% & 76.41\% & -5.74\% \\ \hline
        9 & Y & 58.74\% & -7.96\% & 73.54\% & -4.69\% & 78.51\% & -3.15\% \\ \hline
        6 & N & 42.79\% & -32.95\% & 59.97\% & -22.28\% & 66.63\% & -17.80\% \\ \hline
        6 & Y & 53.51\% & -16.15\% & 69.87\% & -9.45\% & 75.03\% & -7.44\% \\ \hline
        3 & N & 18.67\% & -70.75\% & 34.42\% & -55.39\% & 42.02\% & -48.16\% \\ \hline
        3 & Y & 47.62\% & -25.38\% & 64.37\% & -16.58\% & 69.89\% & -13.78\% \\ \hline
        2 & N & 20.82\% & -67.38\% & 37.01\% & -52.03\% & 45.01\% & -44.47\% \\ \hline
        2 & Y & 46.60\% & -26.98\% & 63.72\% & -17.42\% & 69.53\% & -14.22\% \\ \hline
        1 & N & 5.30\% & -91.70\% & 11.66\% & -84.89\% & 15.88\% & -80.41\% \\ \hline
        1 & Y & 34.72\% & -45.60\% & 51.39\% & -33.40\% & 58.01\% & -28.44\% \\ \hline
    \end{tabular}
    \caption{Impact of structural pruning with and without KALE on the SQUAD retrieval dataset}
    \label{tab:kale-squad}
\end{table*}
\subsection{KALE and Asymmetric Training}
\label{sec:kale-asym-full}
Building on the impact of asymmetry and KALE, we explore comparing them across various datasets as shown in \ref{tab:kale+asym-msmarco}, \ref{tab:kale+asym+nq},\ref{tab:kale+asym-trivia}, \ref{tab:kale+asym-squad}, \ref{tab:kale+asym-scifacts}. 
\begin{table*}[!ht]
    \centering
    \begin{tabular}{|l|l|l|l|l|l|l|l|}
    \hline
        Model & Layers & KALE & MRR@10 & Impact & Top 20 & Impact & Top 100 \\ \hline
        BERT-base & 12 & N & 32.47\% & 0.00\% & 70.47\% & 0.00\% & 88.77\% \\ \hline
        BERT-base & 6 & Y & 28.71\% & -11.57\% & 65.44\% & -7.14\% & 84.68\% \\ \hline
        $6_{kd}-6_{kd}$ & 6 & N & 32.21\% & -0.78\% & 69.94\% & -0.75\% & 88.19\% \\ \hline
        $6_{db}-6_{db}$ & 6 & N & 32.13\% & -1.02\% & 70.37\% & -0.14\% & 88.35\% \\ \hline
        $6_{kd}-3_{kd}$ & 6 & N & 30.44\% & -6.24\% & 67.82\% & -3.76\% & 86.50\% \\ \hline
        BERT-base & 3 & Y & 26.56\% & -18.19\% & 62.36\% & -11.51\% & 82.11\% \\ \hline
        $3_{kd}-3_{kd}$ & 3 & N & 30.01\% & -7.56\% & 67.42\% & -4.33\% & 86.13\% \\ \hline
        $3_{kd}-6_{kd}$ & 3 & N & 29.60\% & -8.82\% & 66.53\% & -5.59\% & 84.79\% \\ \hline
        $6_{kd}-3_{kd}$ & 3 & Y & 28.19\% & -13.16\% & 64.00\% & -9.19\% & 82.95\% \\ \hline
        $6_{kd}-6_{kd}$ & 3 & Y & 30.40\% & -6.37\% & 67.62\% & -4.05\% & 86.75\% \\ \hline
        BERT-base & 2 & Y & 26.10\% & -19.61\% & 61.68\% & -12.48\% & 81.96\% \\ \hline
        $3_{kd}-3_{kd}$ & 2 & Y & 28.57\% & -12.00\% & 65.67\% & -6.81\% & 84.23\% \\ \hline
        $3_{kd}-6_{kd}$ & 2 & Y & 29.52\% & -9.09\% & 66.16\% & -6.12\% & 85.57\% \\ \hline
        $6_{kd}-3_{kd}$ & 2 & Y & 28.07\% & -13.54\% & 64.28\% & -8.78\% & 83.24\% \\ \hline
        $6_{kd}-6_{kd}$ & 2 & Y & 30.00\% & -7.58\% & 66.91\% & -5.06\% & 85.77\% \\ \hline
        BERT-base & 1 & Y & 10.87\% & -66.53\% & 29.80\% & -57.71\% & 48.05\% \\ \hline
        $3_{kd}-3_{kd}$ & 1 & Y & 19.09\% & -41.21\% & 47.56\% & -32.51\% & 66.69\% \\ \hline
        $3_{kd}-6_{kd}$ & 1 & Y & 21.74\% & -33.04\% & 52.29\% & -25.80\% & 72.13\% \\ \hline
        $6_{kd}-3_{kd}$ & 1 & Y & 20.82\% & -35.88\% & 50.92\% & -27.75\% & 71.26\% \\ \hline
        $6_{kd}-6_{kd}$ & 1 & Y & 20.67\% & -36.33\% & 51.81\% & -26.49\% & 70.70\% \\ \hline
    \end{tabular}
    \caption{Impact of model asymmetry and use of KALE for structural pruning on the MSMARCO retrieval dataset }
    \label{tab:kale+asym-msmarco}
\end{table*}

\begin{table*}[!ht]
    \centering
    \begin{tabular}{|l|l|l|l|l|l|l|l|}
    \hline
        Model & Layers & KALE & recall 20 & Impact & recall 100 & Impact & recall 200 \\ \hline
        BERT-base & 12 & N & 79.86\% & 0.00\% & 85.84\% & 0.00\% & 88.42\% \\ \hline
        BERT-base & 6 & Y & 75.51\% & -5.45\% & 83.68\% & -2.52\% & 86.18\% \\ \hline
        $6_{kd}-6_{kd}$ & 6 & N & 75.04\% & -6.03\% & 85.15\% & -0.80\% & 87.45\% \\ \hline
        $6_{db}-6_{db}$ & 6 & N & 73.88\% & -7.49\% & 84.74\% & -1.29\% & 87.26\% \\ \hline
        $6_{kd}-3_{kd}$ & 6 & N & 77.40\% & -3.09\% & 85.37\% & -0.54\% & 87.48\% \\ \hline
        BERT-base & 3 & Y & 72.69\% & -8.98\% & 81.14\% & -5.48\% & 84.76\% \\ \hline
        $3_{kd}-3_{kd}$ & 3 & N & 71.88\% & -9.99\% & 83.66\% & -2.54\% & 86.37\% \\ \hline
        $3_{kd}-6_{kd}$ & 3 & N & 77.01\% & -3.57\% & 85.76\% & -0.09\% & 87.42\% \\ \hline
        $6_{kd}-3_{kd}$ & 3 & Y & 74.16\% & -7.14\% & 83.43\% & -2.81\% & 85.62\% \\ \hline
        $6_{kd}-6_{kd}$ & 3 & Y & 69.28\% & -13.25\% & 80.78\% & -5.89\% & 84.10\% \\ \hline
        BERT-base & 2 & Y & 71.83\% & -10.06\% & 81.94\% & -4.54\% & 84.54\% \\ \hline
        $3_{kd}-3_{kd}$ & 2 & Y & 70.08\% & -12.25\% & 82.71\% & -3.65\% & 85.60\% \\ \hline
        $3_{kd}-6_{kd}$ & 2 & Y & 75.40\% & -5.58\% & 84.27\% & -1.83\% & 86.81\% \\ \hline
        $6_{kd}-3_{kd}$ & 2 & Y & 73.49\% & -7.98\% & 83.02\% & -3.29\% & 85.76\% \\ \hline
        $6_{kd}-6_{kd}$ & 2 & Y & 68.42\% & -14.33\% & 80.39\% & -6.35\% & 83.57\% \\ \hline
        BERT-base & 1 & Y & 58.86\% & -26.30\% & 71.33\% & -16.90\% & 75.65\% \\ \hline
        $3_{kd}-3_{kd}$ & 1 & Y & 62.69\% & -21.50\% & 77.17\% & -10.10\% & 81.33\% \\ \hline
        $3_{kd}-6_{kd}$ & 1 & Y & 68.14\% & -14.68\% & 79.81\% & -7.02\% & 82.94\% \\ \hline
        $6_{kd}-3_{kd}$ & 1 & Y & 63.82\% & -20.09\% & 76.57\% & -10.80\% & 80.33\% \\ \hline
        $6_{kd}-6_{kd}$ & 1 & Y & 60.03\% & -24.83\% & 74.71\% & -12.97\% & 78.64\% \\ \hline
    \end{tabular}
    \caption{Impact of model asymmetry and use of KALE for structural pruning on the NQ retrieval dataset}
    \label{tab:kale+asym+nq}
\end{table*}

\begin{table*}[!ht]
    \centering
    \begin{tabular}{|l|l|l|l|l|l|l|l|}
    \hline
        Model & Layers & KALE & recall 20 & Impact & recall 100 & Impact & recall 200 \\ \hline
        BERT-base & 12 & N & 79.43\% & 0.00\% & 85.03\% & 0.00\% & 86.63\% \\ \hline
        BERT-base & 6 & Y & 75.37\% & -5.11\% & 83.01\% & -2.38\% & 85.25\% \\ \hline
        $6_{kd}-6_{kd}$ & 6 & N & 79.44\% & 0.01\% & 84.96\% & -0.08\% & 86.60\% \\ \hline
        $6_{db}-6_{db}$ & 6 & N & 78.96\% & -0.59\% & 84.83\% & -0.23\% & 86.61\% \\ \hline
        $6_{kd}-3_{kd}$ & 6 & N & 77.31\% & -2.67\% & 84.04\% & -1.17\% & 85.62\% \\ \hline
        BERT-base & 3 & Y & 73.17\% & -7.88\% & 81.67\% & -3.95\% & 84.04\% \\ \hline
        $3_{kd}-3_{kd}$ & 3 & N & 77.80\% & -2.05\% & 84.11\% & -1.09\% & 85.96\% \\ \hline
        $3_{kd}-6_{kd}$ & 3 & N & 77.52\% & -2.40\% & 83.91\% & -1.31\% & 85.72\% \\ \hline
        $6_{kd}-3_{kd}$ & 3 & Y & 74.98\% & -5.60\% & 82.33\% & -3.18\% & 84.35\% \\ \hline
        $6_{kd}-6_{kd}$ & 3 & Y & 76.76\% & -3.36\% & 83.48\% & -1.82\% & 85.40\% \\ \hline
        BERT-base & 2 & Y & 72.39\% & -8.86\% & 81.23\% & -4.47\% & 83.64\% \\ \hline
        $3_{kd}-3_{kd}$ & 2 & Y & 76.48\% & -3.71\% & 83.02\% & -2.36\% & 85.16\% \\ \hline
        $3_{kd}-6_{kd}$ & 2 & Y & 75.98\% & -4.34\% & 82.90\% & -2.50\% & 85.00\% \\ \hline
        $6_{kd}-3_{kd}$ & 2 & Y & 74.60\% & -6.08\% & 82.13\% & -3.41\% & 84.44\% \\ \hline
        $6_{kd}-6_{kd}$ & 2 & Y & 76.56\% & -3.61\% & 83.32\% & -2.01\% & 85.49\% \\ \hline
        BERT-base & 1 & Y & 63.04\% & -20.63\% & 75.40\% & -11.33\% & 79.23\% \\ \hline
        $3_{kd}-3_{kd}$ & 1 & Y & 71.66\% & -9.78\% & 80.82\% & -4.95\% & 83.56\% \\ \hline
        $3_{kd}-6_{kd}$ & 1 & Y & 71.13\% & -10.45\% & 80.23\% & -5.65\% & 82.86\% \\ \hline
        $6_{kd}-3_{kd}$ & 1 & Y & 68.11\% & -14.25\% & 78.65\% & -7.50\% & 81.89\% \\ \hline
        $6_{kd}-6_{kd}$ & 1 & Y & 70.91\% & -10.73\% & 80.31\% & -5.55\% & 83.05\% \\ \hline
    \end{tabular}
    \caption{Impact of model asymmetry and use of KALE for structural pruning on the TriviaQA retrieval dataset}
    \label{tab:kale+asym-trivia}
\end{table*}

\begin{table*}[!ht]
    \centering
    \begin{tabular}{|l|l|l|l|l|l|l|l|}
    \hline
        Model & Layers & KALE & recall 20 & Impact & recall 100 & Impact & recall 200 \\ \hline
        BERT-base & 12 & N & 63.82\% & 0.00\% & 77.16\% & 0.00\% & 81.06\% \\ \hline
        BERT-base & 6 & Y & 53.51\% & -16.15\% & 69.87\% & -9.45\% & 75.03\% \\ \hline
        $6_{kd}-6_{kd}$ & 6 & N & 54.80\% & -14.14\% & 71.94\% & -6.77\% & 77.73\% \\ \hline
        $6_{db}-6_{db}$ & 6 & N & 54.60\% & -14.45\% & 71.69\% & -7.08\% & 77.23\% \\ \hline
        $6_{kd}-3_{kd}$ & 6 & N & 52.97\% & -17.00\% & 70.89\% & -8.13\% & 76.68\% \\ \hline
        BERT-base & 3 & Y & 47.62\% & -25.38\% & 64.37\% & -16.58\% & 69.89\% \\ \hline
        $3_{kd}-3_{kd}$ & 3 & N & 55.05\% & -13.74\% & 71.98\% & -6.72\% & 77.76\% \\ \hline
        $3_{kd}-6_{kd}$ & 3 & N & 48.86\% & -23.43\% & 67.85\% & -12.06\% & 74.04\% \\ \hline
        $6_{kd}-3_{kd}$ & 3 & Y & 44.65\% & -30.04\% & 63.77\% & -17.35\% & 70.79\% \\ \hline
        $6_{kd}-6_{kd}$ & 3 & Y & 45.36\% & -28.93\% & 64.14\% & -16.87\% & 71.07\% \\ \hline
        BERT-base & 2 & Y & 48.43\% & -24.11\% & 67.02\% & -13.14\% & 73.19\% \\ \hline
        $3_{kd}-3_{kd}$ & 2 & Y & 48.43\% & -24.11\% & 67.02\% & -13.14\% & 73.19\% \\ \hline
        $3_{kd}-6_{kd}$ & 2 & Y & 43.45\% & -31.92\% & 62.75\% & -18.68\% & 69.74\% \\ \hline
        $6_{kd}-3_{kd}$ & 2 & Y & 42.90\% & -32.78\% & 62.52\% & -18.97\% & 69.47\% \\ \hline
        $6_{kd}-6_{kd}$ & 2 & Y & 35.08\% & -45.03\% & 52.74\% & -31.65\% & 59.93\% \\ \hline
        BERT-base & 1 & Y & 34.72\% & -45.60\% & 51.39\% & -33.40\% & 58.01\% \\ \hline
        $3_{kd}-3_{kd}$ & 1 & Y & 36.19\% & -43.29\% & 55.62\% & -27.92\% & 62.92\% \\ \hline
        $3_{kd}-6_{kd}$ & 1 & Y & 34.75\% & -45.55\% & 52.26\% & -32.27\% & 59.35\% \\ \hline
        $6_{kd}-3_{kd}$ & 1 & Y & 32.18\% & -49.58\% & 50.88\% & -34.06\% & 58.52\% \\ \hline
        $6_{kd}-6_{kd}$ & 1 & Y & 35.08\% & -45.03\% & 52.74\% & -31.65\% & 59.93\% \\ \hline
    \end{tabular}
    \caption{Impact of model asymmetry and use of KALE for structural pruning on the SQUAD retrieval dataset}
    \label{tab:kale+asym-squad}
\end{table*}

\begin{table*}[!ht]
    \centering
    \begin{tabular}{|l|l|l|l|l|l|l|l|}
    \hline
        Model & Layers & KALE & recip\_rank & Impact & NDC@10 & Impact & Recall 20 \\ \hline
        BERT-base & 12 & N & 59.11\% & 0.00\% & 62.55\% & 0.00\% & 82.38\% \\ \hline
        BERT-base & 6 & Y & 54.99\% & -6.97\% & 58.22\% & -6.92\% & 77.07\% \\ \hline
        $6_{kd}-6_{kd}$ & 6 & N & 65.52\% & 10.84\% & 67.87\% & 8.51\% & 83.92\% \\ \hline
        $6_{db}-6_{db}$ & 6 & N & 66.25\% & 12.08\% & 67.81\% & 8.41\% & 82.16\% \\ \hline
        $6_{kd}-3_{kd}$ & 6 & N & 61.90\% & 4.72\% & 65.30\% & 4.40\% & 82.48\% \\ \hline
        BERT-base & 3 & Y & 55.18\% & -6.65\% & 58.30\% & -6.79\% & 76.73\% \\ \hline
        $3_{kd}-3_{kd}$ & 3 & N & 65.32\% & 10.51\% & 67.51\% & 7.93\% & 84.36\% \\ \hline
        $3_{kd}-6_{kd}$ & 3 & N & 62.78\% & 6.21\% & 64.86\% & 3.69\% & 79.80\% \\ \hline
        $6_{kd}-3_{kd}$ & 3 & Y & 62.07\% & 5.01\% & 64.73\% & 3.49\% & 82.57\% \\ \hline
        $6_{kd}-6_{kd}$ & 3 & Y & 61.82\% & 4.58\% & 65.41\% & 4.57\% & 82.41\% \\ \hline
        BERT-base & 2 & Y & 54.45\% & -7.88\% & 57.71\% & -7.74\% & 76.07\% \\ \hline
        $3_{kd}-3_{kd}$ & 2 & Y & 61.78\% & 4.52\% & 64.78\% & 3.57\% & 82.76\% \\ \hline
        $3_{kd}-6_{kd}$ & 2 & Y & 61.41\% & 3.89\% & 63.61\% & 1.69\% & 82.46\% \\ \hline
        $6_{kd}-3_{kd}$ & 2 & Y & 61.82\% & 4.58\% & 64.80\% & 3.60\% & 82.51\% \\ \hline
        $6_{kd}-6_{kd}$ & 2 & Y & 62.09\% & 5.04\% & 65.27\% & 4.35\% & 81.51\% \\ \hline
        BERT-base & 1 & Y & 40.52\% & -31.45\% & 43.23\% & -30.89\% & 59.00\% \\ \hline
        $3_{kd}-3_{kd}$ & 1 & Y & 42.93\% & -27.37\% & 44.19\% & -29.35\% & 61.06\% \\ \hline
        $3_{kd}-6_{kd}$ & 1 & Y & 42.33\% & -28.39\% & 44.03\% & -29.61\% & 63.33\% \\ \hline
        $6_{kd}-3_{kd}$ & 1 & Y & 42.72\% & -27.73\% & 45.68\% & -26.97\% & 65.81\% \\ \hline
        $6_{kd}-6_{kd}$ & 1 & Y & 45.60\% & -22.86\% & 48.83\% & -21.93\% & 69.11\% \\ \hline
    \end{tabular}
    \caption{Impact of model asymmetry and use of KALE for structural pruning on the SCIFACTS retrieval dataset}
    \label{tab:kale+asym-scifacts}
\end{table*}
\subsection{Inference Benchmarks}
\label{sec:inference-benchmarks}
Evaluation of inference on GPU can be found in \ref{tab:benchmark-gpu-12layer},\ref{tab:benchmark-gpu-9layer},\ref{tab:benchmark-gpu-6layer},\ref{tab:benchmark-gpu-3layer} ,\ref{tab:benchmark-gpu-2layer},\ref{tab:benchmark-gpu-1layer} while CPU results can be found in \ref{tab:benchmark-cpu-12layer}, \ref{tab:benchmark-cpu-9layer}, \ref{tab:benchmark-cpu-6layer}, \ref{tab:benchmark-cpu-3layer}, \ref{tab:benchmark-cpu-2layer}, \ref{tab:benchmark-cpu-1layer}.  

\begin{table*}[!ht]
    \centering
    \begin{tabular}{|l|l|l|l|l|l|l|l|}
    \hline
          & items/sec & Full Time & Mean Time & 95th & 50th & 5th & 99th \\ \hline
        Run 1 & 44.890 & 80.414 & 2.17E-02 & 2.92E-02 & 2.09E-02 & 1.97E-02 & 3.07E-02 \\ \hline
        Run 2 & 48.370 & 74.628 & 2.01E-02 & 2.11E-02 & 2.00E-02 & 1.96E-02 & 2.22E-02 \\ \hline
        Run 3 & 47.290 & 76.334 & 2.06E-02 & 2.19E-02 & 2.04E-02 & 1.96E-02 & 2.28E-02 \\ \hline
        Run 4 & 48.260 & 74.810 & 2.01E-02 & 2.13E-02 & 2.00E-02 & 1.95E-02 & 2.22E-02 \\ \hline
        Run 5 & 47.580 & 75.872 & 2.04E-02 & 2.14E-02 & 2.03E-02 & 1.98E-02 & 2.28E-02 \\ \hline
        average & 47.278 & 76.412 & 2.06E-02 & 2.30E-02 & 2.03E-02 & 1.96E-02 & 2.41E-02 \\ \hline
        stdev & 1.410 & 2.348 & 6.46E-04 & 3.49E-03 & 3.65E-04 & 1.04E-04 & 3.68E-03 \\ \hline
        CI & 1.236 & 2.058 & 5.66E-04 & 3.06E-03 & 3.20E-04 & 9.14E-05 & 3.23E-03 \\ \hline
        Lower & 46.042 & 74.353 & 2.00E-02 & 1.99E-02 & 2.00E-02 & 1.96E-02 & 2.09E-02 \\ \hline
        High & 48.514 & 78.470 & 2.12E-02 & 2.60E-02 & 2.06E-02 & 1.97E-02 & 2.74E-02 \\ \hline
    \end{tabular}
    \caption{Inference Benchmark for 12-layer Query encoder on a CPU using ONNX}
    \label{tab:benchmark-cpu-12layer}
\end{table*}

\begin{table*}[!ht]
    \centering
    \begin{tabular}{|l|l|l|l|l|l|l|l|}
    \hline
        ~ & items/sec & Full Time & Mean Time & 95th & 50th & 5th & 99th \\ \hline
        Run 1 & 63.200 & 57.808 & 1.54E-02 & 1.65E-02 & 1.52E-02 & 1.49E-02 & 2.20E-02 \\ \hline
        Run 2 & 63.570 & 56.787 & 1.52E-02 & 1.60E-02 & 1.50E-02 & 1.48E-02 & 1.70E-02 \\ \hline
        Run 3 & 62.740 & 57.537 & 1.54E-02 & 1.64E-02 & 1.52E-02 & 1.48E-02 & 1.76E-02 \\ \hline
        Run 4 & 63.440 & 56.908 & 1.52E-02 & 1.59E-02 & 1.51E-02 & 1.48E-02 & 1.70E-02 \\ \hline
        Run 5 & 63.250 & 57.077 & 1.53E-02 & 1.60E-02 & 1.51E-02 & 1.48E-02 & 1.69E-02 \\ \hline
        average & 63.240 & 57.223 & 1.53E-02 & 1.62E-02 & 1.51E-02 & 1.48E-02 & 1.81E-02 \\ \hline
        stdev & 0.316 & 0.433 & 1.16E-04 & 2.49E-04 & 6.48E-05 & 6.69E-05 & 2.20E-03 \\ \hline
        CI & 0.277 & 0.380 & 1.02E-04 & 2.18E-04 & 5.68E-05 & 5.86E-05 & 1.93E-03 \\ \hline
        Lower & 62.963 & 56.844 & 1.52E-02 & 1.59E-02 & 1.51E-02 & 1.48E-02 & 1.62E-02 \\ \hline
        High & 63.517 & 57.603 & 1.54E-02 & 1.64E-02 & 1.52E-02 & 1.49E-02 & 2.00E-02 \\ \hline
    \end{tabular}
    \caption{Inference Benchmark for 9-layer Query encoder on a CPU using ONNX}
    \label{tab:benchmark-cpu-9layer}
\end{table*}

\begin{table*}[!ht]
    \centering
    \begin{tabular}{|l|l|l|l|l|l|l|l|}
    \hline
        ~ & items/sec & Full Time & Mean Time & 95th & 50th & 5th & 99th \\ \hline
        Run 1 & 91.090 & 39.631 & 1.04E-02 & 1.11E-02 & 1.03E-02 & 1.02E-02 & 1.19E-02 \\ \hline
        Run 2 & 90.990 & 39.677 & 1.04E-02 & 1.11E-02 & 1.03E-02 & 1.01E-02 & 1.22E-02 \\ \hline
        Run 3 & 91.290 & 39.547 & 1.04E-02 & 1.11E-02 & 1.03E-02 & 1.01E-02 & 1.22E-02 \\ \hline
        Run 4 & 89.420 & 40.372 & 1.06E-02 & 1.24E-02 & 1.02E-02 & 1.01E-02 & 1.51E-02 \\ \hline
        Run 5 & 89.140 & 40.499 & 1.07E-02 & 1.21E-02 & 1.03E-02 & 1.01E-02 & 1.49E-02 \\ \hline
        average & 90.386 & 39.945 & 1.05E-02 & 1.16E-02 & 1.03E-02 & 1.01E-02 & 1.32E-02 \\ \hline
        stdev & 1.020 & 0.452 & 1.23E-04 & 6.03E-04 & 3.95E-05 & 4.27E-05 & 1.61E-03 \\ \hline
        CI & 0.894 & 0.396 & 1.08E-04 & 5.29E-04 & 3.47E-05 & 3.74E-05 & 1.41E-03 \\ \hline
        Lower & 89.492 & 39.549 & 1.04E-02 & 1.10E-02 & 1.03E-02 & 1.01E-02 & 1.18E-02 \\ \hline
        High & 91.280 & 40.342 & 1.06E-02 & 1.21E-02 & 1.03E-02 & 1.02E-02 & 1.47E-02 \\ \hline
    \end{tabular}
    \caption{Inference Benchmark for 6-layer Query encoder on a CPU using ONNX}
    \label{tab:benchmark-cpu-6layer}
\end{table*}

\begin{table*}[!ht]
    \centering
    \begin{tabular}{|l|l|l|l|l|l|l|l|}
    \hline
        ~ & items/sec & Full Time & Mean Time & 95th & 50th & 5th & 99th \\ \hline
        Run 1 & 166.340 & 21.704 & 5.47E-03 & 5.84E-03 & 5.40E-03 & 5.35E-03 & 6.34E-03 \\ \hline
        Run 2 & 164.830 & 21.902 & 5.53E-03 & 6.14E-03 & 5.40E-03 & 5.31E-03 & 7.35E-03 \\ \hline
        Run 3 & 167.570 & 21.544 & 5.43E-03 & 5.87E-03 & 5.34E-03 & 5.30E-03 & 6.42E-03 \\ \hline
        Run 4 & 165.370 & 21.830 & 5.51E-03 & 6.11E-03 & 5.39E-03 & 5.30E-03 & 6.96E-03 \\ \hline
        Run 5 & 165.950 & 21.755 & 5.49E-03 & 5.92E-03 & 5.40E-03 & 5.32E-03 & 6.54E-03 \\ \hline
        average & 166.012 & 21.747 & 5.49E-03 & 5.98E-03 & 5.39E-03 & 5.32E-03 & 6.72E-03 \\ \hline
        stdev & 1.043 & 0.136 & 3.58E-05 & 1.41E-04 & 2.49E-05 & 2.20E-05 & 4.23E-04 \\ \hline
        CI & 0.914 & 0.119 & 3.14E-05 & 1.23E-04 & 2.18E-05 & 1.93E-05 & 3.71E-04 \\ \hline
        Lower & 165.098 & 21.628 & 5.45E-03 & 5.86E-03 & 5.37E-03 & 5.30E-03 & 6.35E-03 \\ \hline
        High & 166.926 & 21.867 & 5.52E-03 & 6.10E-03 & 5.41E-03 & 5.33E-03 & 7.09E-03 \\ \hline
        BERT-base & 2 & Y & 54.45\% & -7.88\% & 57.71\% & -7.74\% & 76.07\% \\ \hline
    \end{tabular}
    \caption{Inference Benchmark for 3-layer Query encoder on a CPU using ONNX}
    \label{tab:benchmark-cpu-3layer}
\end{table*}

\begin{table*}[!ht]
    \centering
    \begin{tabular}{|l|l|l|l|l|l|l|l|}
    \hline
        ~ & items/sec & Full Time & Mean Time & 95th & 50th & 5th & 99th \\ \hline
        Run 1 & 228.690 & 15.786 & 3.85E-03 & 4.53E-03 & 3.72E-03 & 3.67E-03 & 5.29E-03 \\ \hline
        Run 2 & 230.420 & 15.668 & 3.81E-03 & 4.24E-03 & 3.74E-03 & 3.65E-03 & 4.72E-03 \\ \hline
        Run 3 & 228.800 & 15.779 & 3.84E-03 & 4.23E-03 & 3.77E-03 & 3.73E-03 & 4.68E-03 \\ \hline
        Run 4 & 230.530 & 15.661 & 3.81E-03 & 4.23E-03 & 3.74E-03 & 3.68E-03 & 4.63E-03 \\ \hline
        Run 5 & 229.890 & 15.704 & 3.82E-03 & 4.25E-03 & 3.75E-03 & 3.70E-03 & 4.64E-03 \\ \hline
        average & 229.666 & 15.720 & 3.83E-03 & 4.29E-03 & 3.74E-03 & 3.69E-03 & 4.79E-03 \\ \hline
        stdev & 0.876 & 0.060 & 1.72E-05 & 1.32E-04 & 1.84E-05 & 3.00E-05 & 2.81E-04 \\ \hline
        CI & 0.768 & 0.053 & 1.51E-05 & 1.16E-04 & 1.61E-05 & 2.63E-05 & 2.47E-04 \\ \hline
        Lower & 228.898 & 15.667 & 3.81E-03 & 4.18E-03 & 3.73E-03 & 3.66E-03 & 4.55E-03 \\ \hline
        High & 230.434 & 15.772 & 3.84E-03 & 4.41E-03 & 3.76E-03 & 3.71E-03 & 5.04E-03 \\ \hline
    \end{tabular}
    \caption{Inference Benchmark for 2 layer Query encoder on a CPU using ONNX}
    \label{tab:benchmark-cpu-2layer}
\end{table*}

\begin{table*}[!ht]
    \centering
    \begin{tabular}{|l|l|l|l|l|l|l|l|}
    \hline
        ~ & items/sec & Full Time & Mean Time & 95th & 50th & 5th & 99th \\ \hline
        Run 1 & 378.680 & 9.534 & 2.14E-03 & 2.39E-03 & 2.10E-03 & 2.08E-03 & 2.88E-03 \\ \hline
        Run 2 & 378.950 & 9.528 & 2.14E-03 & 2.31E-03 & 2.11E-03 & 2.08E-03 & 2.66E-03 \\ \hline
        Run 3 & 377.750 & 9.558 & 2.13E-03 & 2.30E-03 & 2.12E-03 & 2.06E-03 & 2.67E-03 \\ \hline
        Run 4 & 376.560 & 9.588 & 2.16E-03 & 2.35E-03 & 2.12E-03 & 2.06E-03 & 2.74E-03 \\ \hline
        Run 5 & 380.730 & 9.483 & 2.14E-03 & 2.30E-03 & 2.11E-03 & 2.08E-03 & 2.66E-03 \\ \hline
        average & 378.534 & 9.538 & 2.15E-03 & 2.33E-03 & 2.11E-03 & 2.07E-03 & 2.72E-03 \\ \hline
        stdev & 1.543 & 0.039 & 7.46E-06 & 3.64E-05 & 8.72E-06 & 9.49E-06 & 9.64E-05 \\ \hline
        CI & 1.353 & 0.034 & 6.54E-06 & 3.19E-05 & 7.65E-06 & 8.31E-06 & 8.45E-05 \\ \hline
        Lower & 377.181 & 9.504 & 2.14E-03 & 2.30E-03 & 2.11E-03 & 2.06E-03 & 2.64E-03 \\ \hline
        High & 379.887 & 9.572 & 2.15E-03 & 2.36E-03 & 2.12E-03 & 2.08E-03 & 2.81E-03 \\ \hline
    \end{tabular}
    \caption{Inference Benchmark for 1 layer Query encoder on a CPU using ONNX}
    \label{tab:benchmark-cpu-1layer}
\end{table*}

\begin{table*}[!ht]
    \centering
    \begin{tabular}{|l|l|l|l|l|l|l|l|}
    \hline
        ~ & items/sec & Full Time & Mean Time & 95th & 50th & 5th & 99th \\ \hline
        Run 1 & 103.16 & 35.00 & 9.22E-03 & 9.33E-03 & 9.16E-03 & 9.08E-03 & 1.20E-02 \\ \hline
        Run 2 & 111.51 & 32.36 & 8.50E-03 & 8.61E-03 & 8.47E-03 & 8.42E-03 & 8.73E-03 \\ \hline
        Run 3 & 114.02 & 31.66 & 8.31E-03 & 8.41E-03 & 8.28E-03 & 8.22E-03 & 8.60E-03 \\ \hline
        Run 4 & 90.39 & 39.94 & 1.06E-02 & 1.07E-02 & 1.05E-02 & 1.04E-02 & 1.25E-02 \\ \hline
        Run 5 & 110.18 & 32.77 & 8.62E-03 & 8.74E-03 & 8.58E-03 & 8.51E-03 & 9.06E-03 \\ \hline
        average & 105.85 & 34.35 & 9.04E-03 & 9.15E-03 & 9.00E-03 & 8.93E-03 & 1.02E-02 \\ \hline
        stdev & 9.54 & 3.37 & 9.17E-04 & 9.19E-04 & 9.04E-04 & 9.02E-04 & 1.92E-03 \\ \hline
        CI & 8.36 & 2.95 & 8.04E-04 & 8.06E-04 & 7.92E-04 & 7.91E-04 & 1.68E-03 \\ \hline
        Lower & 97.49 & 31.40 & 8.24E-03 & 8.35E-03 & 8.21E-03 & 8.14E-03 & 8.50E-03 \\ \hline
        High & 114.21 & 37.30 & 9.85E-03 & 9.96E-03 & 9.79E-03 & 9.73E-03 & 1.19E-02 \\ \hline
    \end{tabular}
    \caption{Inference Benchmark for 12-layer Query encoder on a T4 GPU}
    \label{tab:benchmark-gpu-12layer}
\end{table*}

\begin{table*}[!ht]
    \centering
    \begin{tabular}{|l|l|l|l|l|l|l|l|}
    \hline
        ~ & items/sec & Full Time & Mean Time & 95th & 50th & 5th & 99th \\ \hline
        Run 1 & 140.35 & 25.72 & 6.69E-03 & 6.78E-03 & 6.66E-03 & 6.61E-03 & 6.94E-03 \\ \hline
        Run 2 & 148.25 & 24.35 & 6.31E-03 & 6.52E-03 & 6.26E-03 & 6.22E-03 & 6.64E-03 \\ \hline
        Run 3 & 147.04 & 24.55 & 6.37E-03 & 6.47E-03 & 6.32E-03 & 6.28E-03 & 7.19E-03 \\ \hline
        Run 4 & 116.15 & 31.08 & 8.14E-03 & 8.25E-03 & 8.09E-03 & 8.01E-03 & 1.09E-02 \\ \hline
        Run 5 & 145.68 & 24.78 & 6.44E-03 & 6.50E-03 & 6.39E-03 & 6.35E-03 & 8.83E-03 \\ \hline
        average & 139.49 & 26.10 & 6.79E-03 & 6.91E-03 & 6.74E-03 & 6.69E-03 & 8.11E-03 \\ \hline
        stdev & 13.39 & 2.84 & 7.70E-04 & 7.62E-04 & 7.66E-04 & 7.52E-04 & 1.79E-03 \\ \hline
        CI & 11.74 & 2.49 & 6.75E-04 & 6.68E-04 & 6.72E-04 & 6.59E-04 & 1.57E-03 \\ \hline
        Lower & 127.75 & 23.61 & 6.11E-03 & 6.24E-03 & 6.07E-03 & 6.04E-03 & 6.54E-03 \\ \hline
        High & 151.23 & 28.58 & 7.46E-03 & 7.57E-03 & 7.42E-03 & 7.35E-03 & 9.67E-03 \\ \hline
    \end{tabular}
    \caption{Inference Benchmark for 9-layer Query encoder on a T4 GPU}
    \label{tab:benchmark-gpu-9layer}
\end{table*}

\begin{table*}[!ht]
    \centering
    \begin{tabular}{|l|l|l|l|l|l|l|l|}
    \hline
        ~ & items/sec & Full Time & Mean Time & 95th & 50th & 5th & 99th \\ \hline
        Run 1 & 163.72 & 22.05 & 5.67E-03 & 5.75E-03 & 5.62E-03 & 5.56E-03 & 7.75E-03 \\ \hline
        Run 2 & 161.90 & 22.30 & 5.74E-03 & 5.81E-03 & 5.70E-03 & 5.63E-03 & 6.17E-03 \\ \hline
        Run 3 & 165.07 & 21.87 & 5.62E-03 & 5.70E-03 & 5.58E-03 & 5.51E-03 & 6.86E-03 \\ \hline
        Run 4 & 189.71 & 19.03 & 4.84E-03 & 4.92E-03 & 4.82E-03 & 4.77E-03 & 5.07E-03 \\ \hline
        Run 5 & 181.29 & 19.91 & 5.07E-03 & 5.92E-03 & 4.94E-03 & 4.88E-03 & 6.68E-03 \\ \hline
        average & 172.34 & 21.03 & 5.39E-03 & 5.62E-03 & 5.33E-03 & 5.27E-03 & 6.51E-03 \\ \hline
        stdev & 12.43 & 1.47 & 4.07E-04 & 3.99E-04 & 4.17E-04 & 4.11E-04 & 9.85E-04 \\ \hline
        CI & 10.89 & 1.29 & 3.56E-04 & 3.50E-04 & 3.65E-04 & 3.61E-04 & 8.63E-04 \\ \hline
        Lower & 161.44 & 19.75 & 5.03E-03 & 5.27E-03 & 4.97E-03 & 4.91E-03 & 5.64E-03 \\ \hline
        High & 183.23 & 22.32 & 5.74E-03 & 5.97E-03 & 5.70E-03 & 5.63E-03 & 7.37E-03 \\ \hline
    \end{tabular}
    \caption{Inference Benchmark for 6-layer Query encoder on a T4 GPU}
    \label{tab:benchmark-gpu-6layer}
\end{table*}

\begin{table*}[!ht]
    \centering
    \begin{tabular}{|l|l|l|l|l|l|l|l|}
    \hline
        ~ & items/sec & Full Time & Mean Time & 95th & 50th & 5th & 99th \\ \hline
        Run 1 & 269.73 & 13.39 & 3.28E-03 & 3.30E-03 & 3.26E-03 & 3.20E-03 & 3.87E-03 \\ \hline
        Run 2 & 282.90 & 12.76 & 3.12E-03 & 3.38E-03 & 3.23E-03 & 2.65E-03 & 4.39E-03 \\ \hline
        Run 3 & 268.47 & 13.45 & 3.30E-03 & 3.31E-03 & 3.28E-03 & 3.25E-03 & 3.76E-03 \\ \hline
        Run 4 & 318.47 & 11.34 & 2.74E-03 & 2.79E-03 & 2.72E-03 & 2.69E-03 & 3.17E-03 \\ \hline
        Run 5 & 357.68 & 10.09 & 2.43E-03 & 2.50E-03 & 2.41E-03 & 2.39E-03 & 2.69E-03 \\ \hline
        average & 299.45 & 12.21 & 2.97E-03 & 3.05E-03 & 2.98E-03 & 2.84E-03 & 3.58E-03 \\ \hline
        stdev & 38.31 & 1.45 & 3.78E-04 & 3.90E-04 & 3.93E-04 & 3.75E-04 & 6.58E-04 \\ \hline
        CI & 33.58 & 1.27 & 3.31E-04 & 3.42E-04 & 3.45E-04 & 3.29E-04 & 5.77E-04 \\ \hline
        Lower & 265.87 & 10.93 & 2.64E-03 & 2.71E-03 & 2.64E-03 & 2.51E-03 & 3.00E-03 \\ \hline
        High & 333.03 & 13.48 & 3.30E-03 & 3.40E-03 & 3.33E-03 & 3.16E-03 & 4.16E-03 \\ \hline
    \end{tabular}
    \caption{Inference Benchmark for 3-layer Query encoder on a T4 GPU}
    \label{tab:benchmark-gpu-3layer}
\end{table*}

\begin{table*}[!ht]
    \centering
    \begin{tabular}{|l|l|l|l|l|l|l|l|}
    \hline
        ~ & items/sec & Full Time & Mean Time & 95th & 50th & 5th & 99th \\ \hline
        Run 1 & 465.83 & 7.75 & 1.78E-03 & 1.83E-03 & 1.76E-03 & 1.74E-03 & 2.53E-03 \\ \hline
        Run 2 & 435.46 & 8.29 & 1.92E-03 & 2.01E-03 & 1.91E-03 & 1.89E-03 & 2.04E-03 \\ \hline
        Run 3 & 471.01 & 7.67 & 1.77E-03 & 1.84E-03 & 1.75E-03 & 1.74E-03 & 1.95E-03 \\ \hline
        Run 4 & 413.49 & 8.73 & 2.02E-03 & 2.06E-03 & 2.00E-03 & 1.96E-03 & 2.61E-03 \\ \hline
        Run 5 & 421.32 & 8.57 & 1.98E-03 & 2.05E-03 & 1.96E-03 & 1.94E-03 & 2.07E-03 \\ \hline
        average & 441.42 & 8.20 & 1.89E-03 & 1.96E-03 & 1.88E-03 & 1.86E-03 & 2.24E-03 \\ \hline
        stdev & 25.94 & 0.48 & 1.15E-04 & 1.12E-04 & 1.15E-04 & 1.07E-04 & 3.07E-04 \\ \hline
        CI & 22.73 & 0.42 & 1.00E-04 & 9.83E-05 & 1.01E-04 & 9.34E-05 & 2.69E-04 \\ \hline
        Lower & 418.69 & 7.78 & 1.79E-03 & 1.86E-03 & 1.78E-03 & 1.76E-03 & 1.97E-03 \\ \hline
        High & 464.16 & 8.62 & 1.99E-03 & 2.05E-03 & 1.98E-03 & 1.95E-03 & 2.51E-03 \\ \hline
    \end{tabular}
    \caption{Inference Benchmark for 2-layer Query encoder on a T4 GPU}
    \label{tab:benchmark-gpu-2layer}
\end{table*}

\begin{table*}[!ht]
    \centering
    \begin{tabular}{|l|l|l|l|l|l|l|l|}
    \hline
        ~ & items/sec & Full Time & Mean Time & 95th & 50th & 5th & 99th \\ \hline
        Run 1 & 627.64 & 5.75 & 1.22E-03 & 1.26E-03 & 1.21E-03 & 1.20E-03 & 1.28E-03 \\ \hline
        Run 2 & 673.96 & 5.36 & 1.13E-03 & 1.18E-03 & 1.12E-03 & 1.11E-03 & 1.22E-03 \\ \hline
        Run 3 & 651.45 & 5.54 & 1.18E-03 & 1.24E-03 & 1.17E-03 & 1.16E-03 & 1.28E-03 \\ \hline
        Run 4 & 677.99 & 5.33 & 1.12E-03 & 1.19E-03 & 1.11E-03 & 1.10E-03 & 1.22E-03 \\ \hline
        Run 5 & 672.16 & 5.37 & 1.13E-03 & 1.18E-03 & 1.12E-03 & 1.11E-03 & 1.22E-03 \\ \hline
        average & 660.64 & 5.47 & 1.15E-03 & 1.21E-03 & 1.14E-03 & 1.14E-03 & 1.24E-03 \\ \hline
        stdev & 21.12 & 0.18 & 4.28E-05 & 3.74E-05 & 4.44E-05 & 4.25E-05 & 3.30E-05 \\ \hline
        CI & 18.51 & 0.16 & 3.75E-05 & 3.27E-05 & 3.89E-05 & 3.72E-05 & 2.89E-05 \\ \hline
        Lower & 642.13 & 5.31 & 1.12E-03 & 1.18E-03 & 1.11E-03 & 1.10E-03 & 1.21E-03 \\ \hline
        High & 679.15 & 5.63 & 1.19E-03 & 1.24E-03 & 1.18E-03 & 1.17E-03 & 1.27E-03 \\ \hline
    \end{tabular}
    \caption{Inference Benchmark for 1-layer Query encoder on a T4 GPU}
    \label{tab:benchmark-gpu-1layer}
\end{table*}
\end{document}